\documentclass[10pt,twocolumn,letterpaper]{article}

\usepackage[pagenumbers]{iccv} 

\usepackage[utf8]{inputenc}
\usepackage[T1]{fontenc}
\usepackage{xcolor}
\usepackage{colortbl}
\usepackage{booktabs}
\usepackage{mathtools,amsfonts,nicefrac,bm}
\usepackage{microtype}
\usepackage{graphicx}
\usepackage{xparse}
\usepackage{xargs}
\usepackage{etoolbox}
\usepackage[english]{babel}
\usepackage{lipsum}
\usepackage{algorithm}
\usepackage{algpseudocode}
\usepackage{enumitem}
\usepackage{mleftright}
\usepackage{multirow}
\usepackage{float}
\usepackage{stfloats}

\mleftright

\usepackage{siunitx}
\usepackage{tikz}
\usepackage{twemojis}
\usepackage{pgfplots,pgfplotstable}
\pgfplotsset{
    compat=1.17,
    compat/show suggested version=false,
    colormap={diverging}{
            rgb255=(215,25,28)
            rgb255=(253,174,97)
            rgb255=(255,255,191)
            rgb255=(171,221,164)
            rgb255=(43,131,186)
        },
    colormap={sequential}{
            rgb255=(255,255,204)
            rgb255=(161,218,180)
            rgb255=(65,182,196)
            rgb255=(44,127,184)
            rgb255=(37,52,148)
        },
    legend image code/.code={
            \draw[mark indices={2}] plot coordinates {
                    (0cm,0cm) (0.2cm,0cm) (0.4cm,0cm)
                };
        }
}
\usepgfplotslibrary{fillbetween}
\usetikzlibrary{backgrounds,positioning,arrows.meta,fit,calc}

\pgfdeclarelayer{-1}
\pgfsetlayers{-1,main}
\tikzset{zlevel/.style={%
        execute at begin scope={\pgfonlayer{#1}},
        execute at end scope={\endpgfonlayer}
}}

\usepackage{acro}
\DeclareAcronym{cfg}{
    short=CFG,
    long=classifier-free guidance
}

\DeclareAcronym{nfe}{
    short=NFE,
    long=neural function evaluation
}

\DeclareAcronym{mlp}{
    short=MLP,
    long=multi-layer perceptron
}

\DeclareAcronym{agd}{
    short=AGD,
    long=adapter guidance distillation
}

\DeclareAcronym{gd}{
    short=GD,
    long=guidance distillation
}

\DeclareAcronym{dit}{
    short=DiT,
    long=Diffusion Transformer
}

\DeclareAcronym{sd2}{
    short=SD2.1,
    long=Stable Diffusion 2.1
}

\DeclareAcronym{sdxl}{
    short=SDXL,
    long=Stable Diffusion XL
}

\DeclareAcronym{ddpm}{
    short=DDPM,
    long=denoising diffusion probabilistic models
}

\DeclareAcronym{sde}{
    short=SDE,
    long=stochastic differential equation
}

\DeclareAcronym{ode}{
    short=ODE,
    long=ordinary differential equation
}

\DeclareAcronym{fid}{
    short=FID,
    long=Fréchet inception distance
}

\definecolor{black}{HTML}{100f0f}
\definecolor{C0}{HTML}{4385be}
\definecolor{C1}{HTML}{d14d41}
\definecolor{C2}{HTML}{879a39}
\definecolor{C3}{HTML}{ce5d97}
\definecolor{C4}{HTML}{d0a215}
\definecolor{C5}{HTML}{3aa99f}
\definecolor{C6}{HTML}{da702c}
\definecolor{C7}{HTML}{8b7ec8}

\definecolor{Blue}{HTML}{C6DDE8}
\definecolor{Green}{HTML}{DDE2B2}
\definecolor{Yellow}{HTML}{F6E2A0}
\definecolor{Cyan}{HTML}{BFE8D9}

\DeclarePairedDelimiter{\abs}{\lvert}{\rvert}

\newcommand{\lft}{\mathopen{}\mathclose\bgroup\left}
\newcommand{\rgt}{\aftergroup\egroup\right}

\newcommand{\pdata}{p_{\textnormal{data}}}
\newcommand{\R}{\mathbb{R}}

\newcommand{\E}{\mathbb{E}}

\renewcommand{\vec}[1]{\bm{#1}}
\newcommand{\mat}[1]{\mathbf{#1}}

\makeatletter
\DeclareRobustCommand\onedot{\futurelet\@let@token\@onedot}
\def\@onedot{\ifx\@let@token.\else.\null\fi\xspace}

\def\eg{{e.g}\onedot} 
\def\ie{{i.e}\onedot} 
 
\def\etc{{etc}\onedot} 
 
\makeatother

\DeclareMathOperator*{\argmin}{argmin} 

\DeclareMathOperator{\grad}{\nabla}  
\newcommand{\dd}{\mathrm{d}}

\newcommand{\prn}[1]{\lft( #1 \rgt)}

\newcommand{\wcfg}{\omega}
\newcommand{\pred}{\vec{\epsilon}_{\vec{\theta}}(\vec{x}_t, t, c)}
\newcommand{\prednull}{\vec{\epsilon}_{\vec{\theta}}(\vec{x}_t, t,\emptyset)}
\newcommand{\predcfg}{\tilde{\vec{\epsilon}}_{\vec{\theta}}(\vec{x}_t, t, c, \omega)}

\newcommand{\denoiser}{D_{\vec{\theta}}(\vec{x}_t, t)}
\newcommand{\adapter}{g_{\vec{\psi}}(\mat{Z}, \omega, t, c)}

\newcommand{\trp}[1]{#1^{\top}}

\renewcommand{\abs}[1]{\left\lvert #1 \right\rvert} %

\definecolor{LightBlue}{HTML}{C6DDE8}
\colorlet{LightGray}{White!50!LightBlue}
\newcommand{\mycc}{\cellcolor{LightGray}}

\definecolor{C0}{rgb}{0.121569, 0.466667, 0.705882}
\definecolor{C1}{rgb}{1.000000, 0.498039, 0.054902}
\definecolor{C2}{rgb}{0.172549, 0.627451, 0.172549}
\definecolor{C3}{rgb}{0.839216, 0.152941, 0.156863}
\definecolor{C4}{rgb}{0.580392, 0.403922, 0.741176}
\definecolor{C5}{rgb}{0.549020, 0.337255, 0.294118}
\definecolor{C6}{rgb}{0.890196, 0.466667, 0.760784}
\definecolor{C7}{rgb}{0.498039, 0.498039, 0.498039}
\definecolor{C8}{rgb}{0.737255, 0.741176, 0.133333}
\definecolor{C9}{rgb}{0.090196, 0.745098, 0.811765}

\newcommand{\Teaser}{%
\centering
\begin{tikzpicture}[every node/.style={inner xsep=0pt,outer xsep=0pt}]
    \node at (0, 0) (img1) {\includegraphics[width=0.4\textwidth]{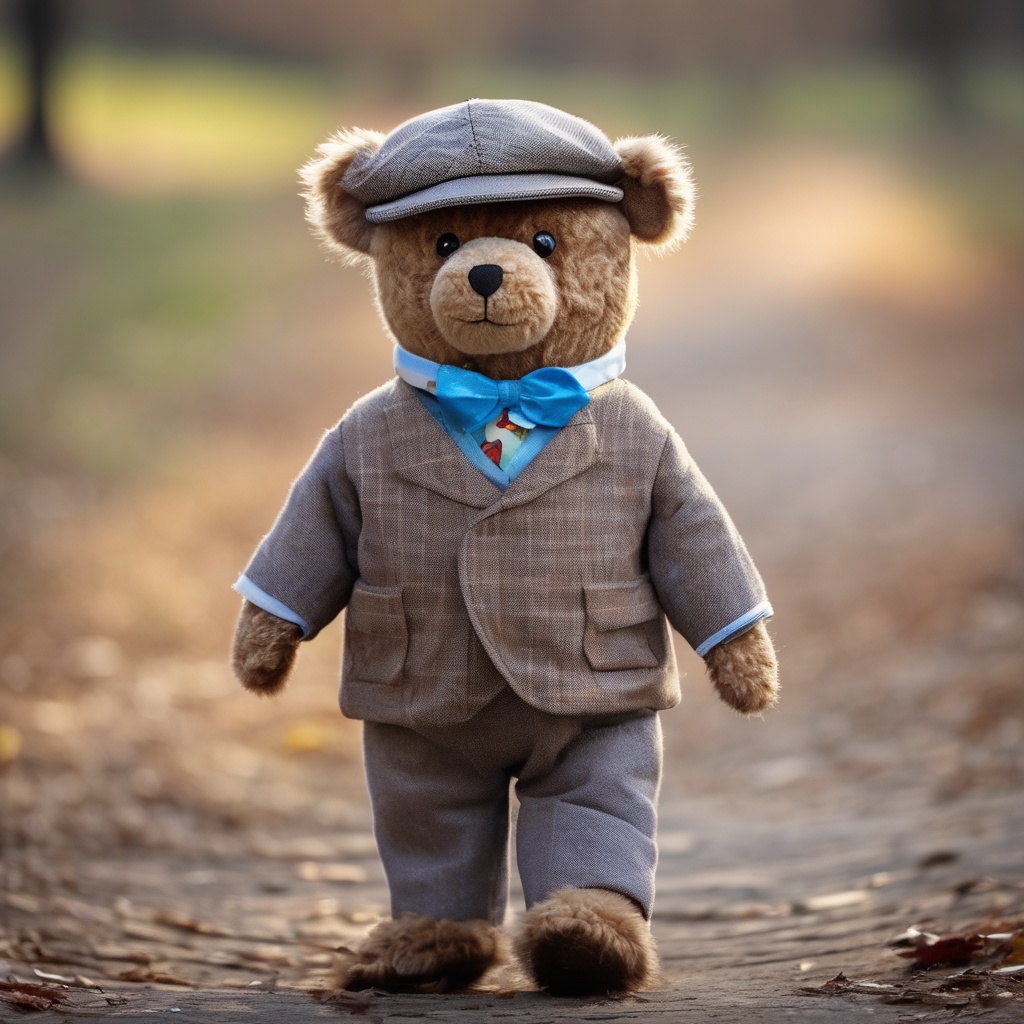}};
    \node [right=0 of img1] (img2) {\includegraphics[width=0.4\textwidth]{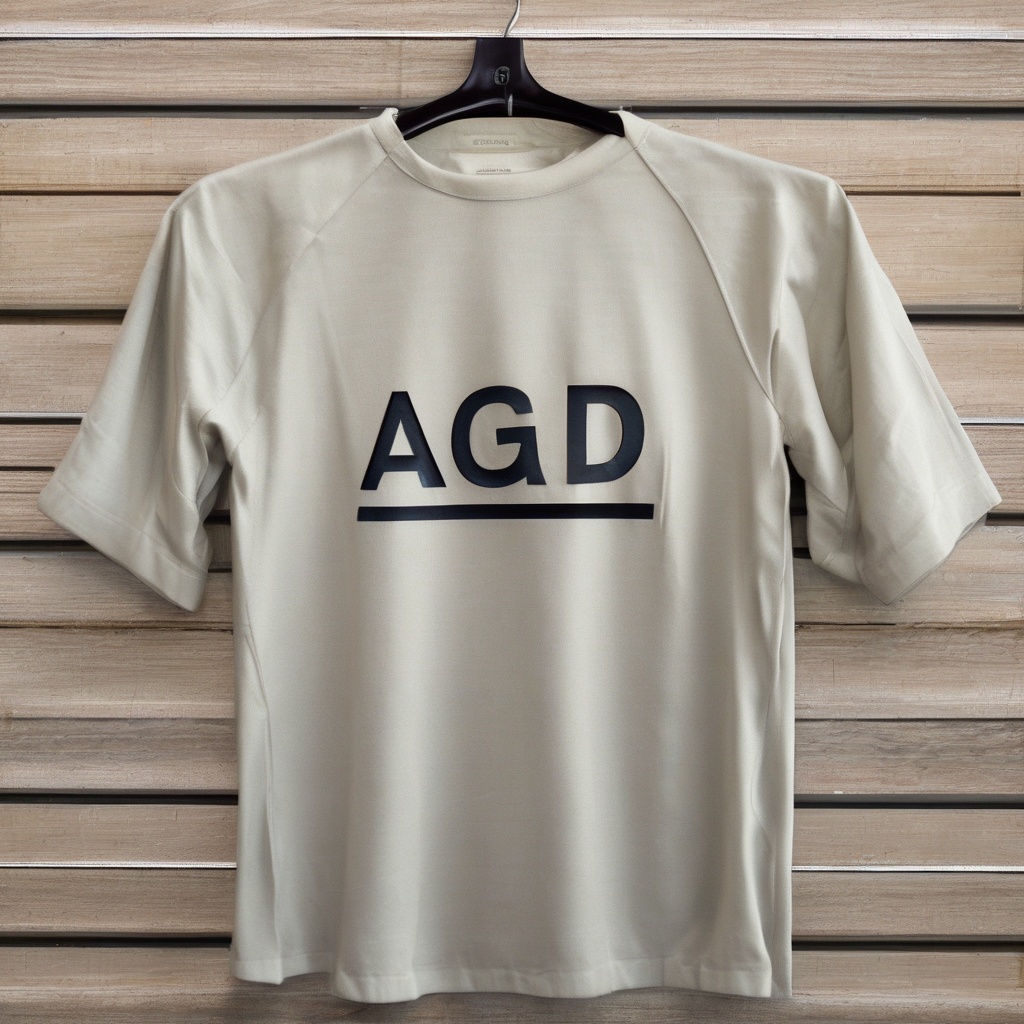}};
    \node [right=0 of img2.north east, anchor=north west] (img3) {\includegraphics[width=0.2\textwidth]{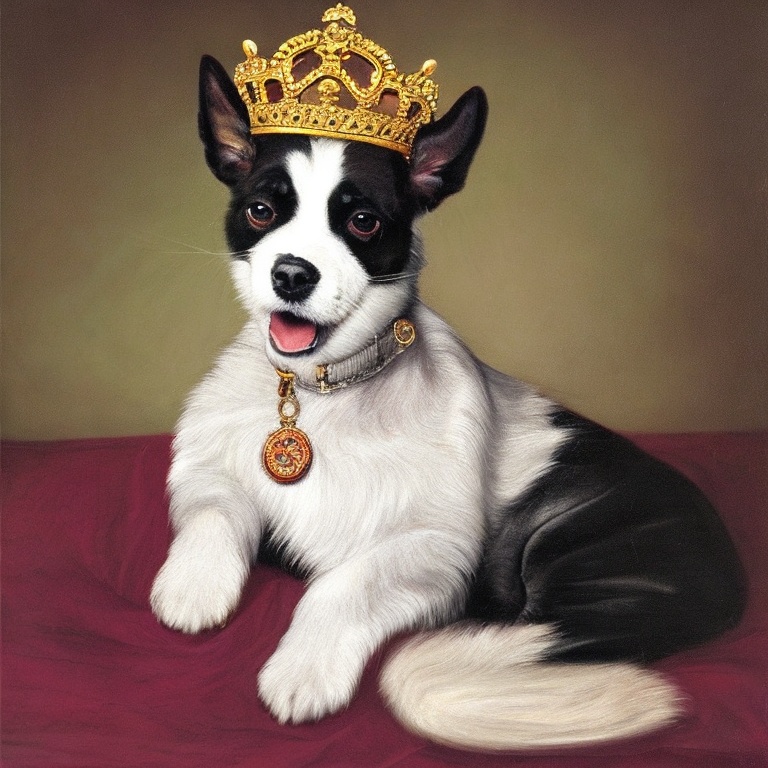}};
    \node [right=0 of img2.south east, anchor=south west] (img4) {\includegraphics[width=0.2\textwidth]{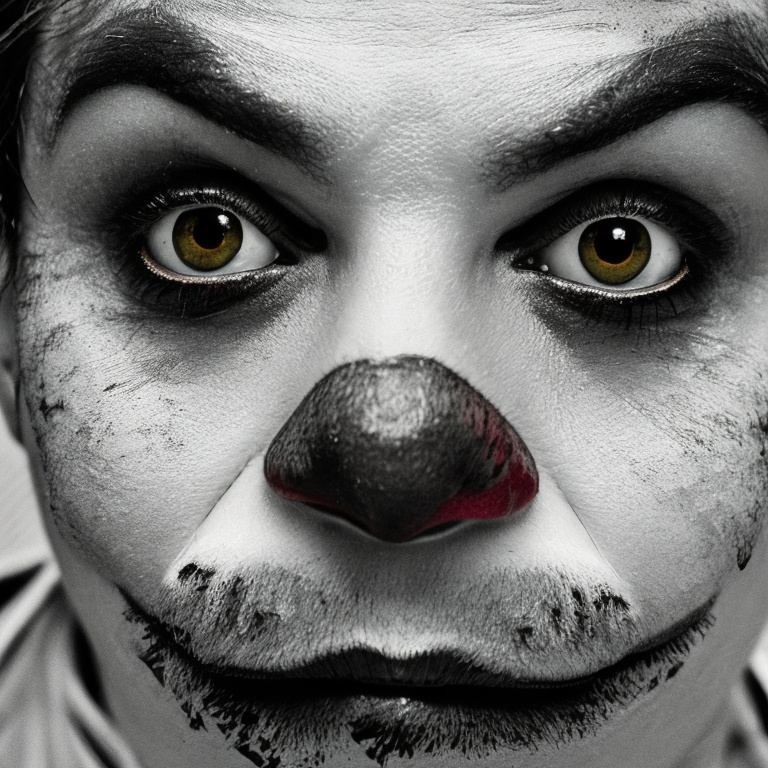}};

    \node [below=0.6 of img1.south west, anchor=north west] (img5) {\includegraphics[width=0.2\textwidth]{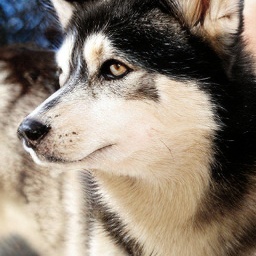}};
    \node [right=0 of img5] (img6) {\includegraphics[width=0.2\textwidth]{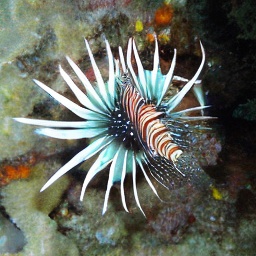}};
    \node [right=0 of img6] (img7) {\includegraphics[width=0.2\textwidth]{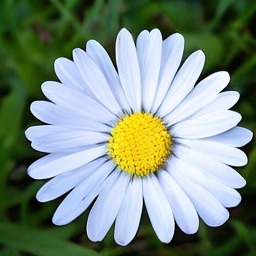}};
    \node [right=0 of img7] (img8) {\includegraphics[width=0.2\textwidth]{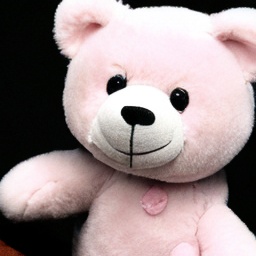}};
    \node [right=0 of img8] (img9) {\includegraphics[width=0.2\textwidth]{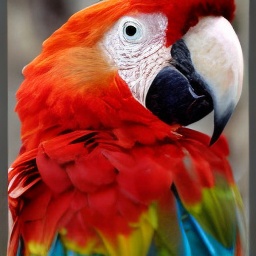}};
    
    \node [below=0 of img1.south east] {\textit{Stable Diffusion XL}};
    \node [below=0 of img4] {\textit{Stable Diffusion 2.1}};
    \node [below=0 of img7] {\textit{Diffusion Transformer}};
\end{tikzpicture}%
\vspace{-.1cm}
\captionof{figure}{Generated samples using \acf{agd} applied to various models. By efficiently baking \acf{cfg} into the base diffusion model, \acs{agd} generates high-quality samples with only a single forward pass per inference step, which results in doubling the sampling speed compared to standard \acs{cfg}.}
}

\newcommand{\TrajectoryDensityTikz}[1]{%
\begin{tikzpicture}[rotate=180]
    \begin{axis}[sharp corners, axis on top, width=0.32\linewidth, height=0.22\linewidth, enlargelimits=false, ticks=none, scale only axis]
        \addplot graphics [xmin=0, xmax=1000, ymin=-2.5, ymax=2.5] {figures/trajectories/#1};
    \end{axis}
\end{tikzpicture}%
}

\newcommand{\TrajectoryDensities}{%
\begin{figure}[t]
    \centering
    \begin{subfigure}{\linewidth}
        \begin{center}
            \begin{tikzpicture}[rotate=90]
                \begin{axis}[axis on top, width=0.3\linewidth, height=0.1\linewidth, scale only axis, ticks=none, enlarge x limits=false, ymin=0.0, ymax=0.6]
                    \addplot [black, thick, mark=none] table [x=x, y=pdf, col sep=comma] {figures/trajectories/prior.csv};
                \end{axis}
            \end{tikzpicture}
            \hfill
            \begin{tikzpicture}[rotate=180]
                \begin{axis}[sharp corners, axis on top, width=0.76\linewidth, height=0.3\linewidth, enlargelimits=false, ticks=none, scale only axis]
                    \addplot graphics [xmin=0, xmax=1000, ymin=-2.5, ymax=2.5] {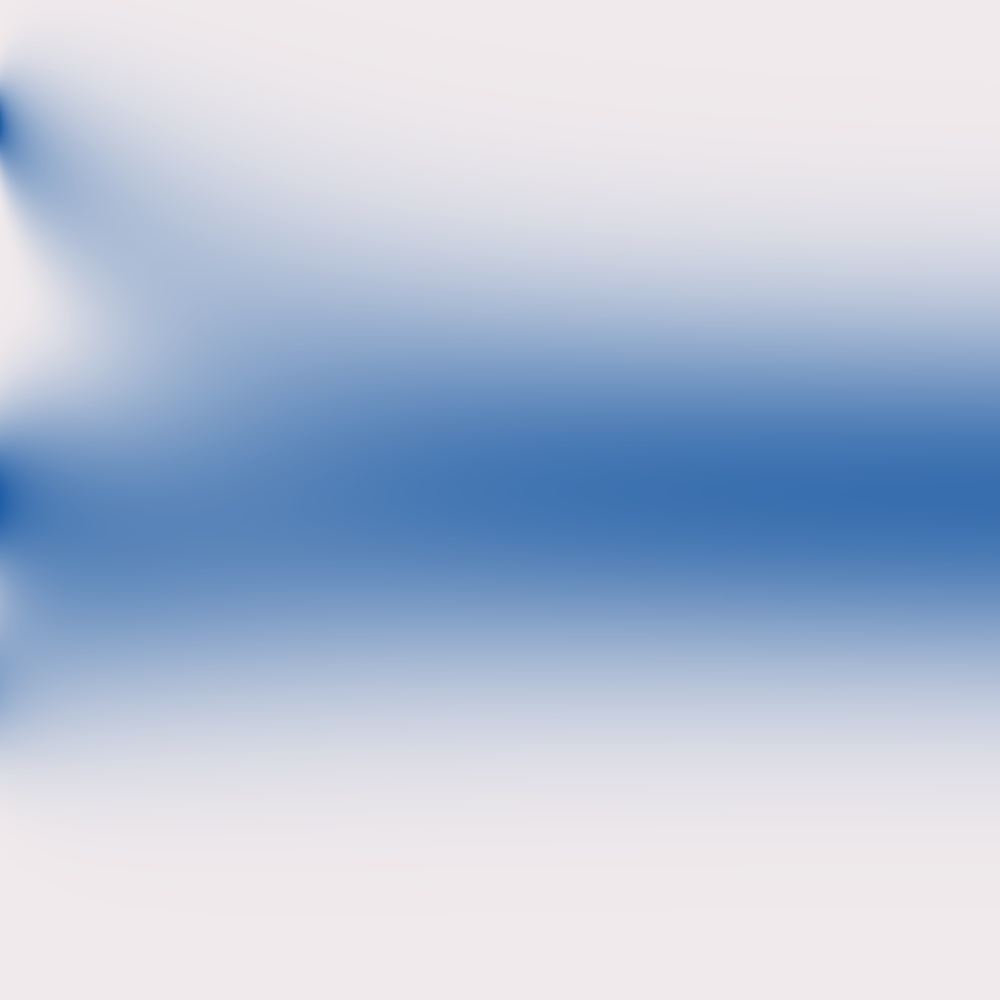};
                \end{axis}
            \end{tikzpicture}
            \hfill
            \begin{tikzpicture}[rotate=-90]
                \begin{axis}[axis on top, width=0.3\linewidth, height=0.1\linewidth, scale only axis, ticks=none, enlarge x limits=false, ymin=0.0, ymax=0.6]
                    \addplot [black, thick, mark=none] table [x=x, y=pdf, col sep=comma] {figures/trajectories/target.csv};
                \end{axis}
            \end{tikzpicture}
        \end{center}
        \caption{Diffusion process with a mixture of Gaussians as the data distribution.}
        \label{fig:trajectory_densities:diffusion_process}
    \end{subfigure}
    \par\medskip
    \begin{subfigure}{\linewidth}
        \begin{center}
            \TrajectoryDensityTikz{diffusion_0} \hfill \TrajectoryDensityTikz{diffusion_1} \hfill \TrajectoryDensityTikz{diffusion_2}
        \end{center}
        \caption{Standard diffusion trajectory densities for each class obtained by adding noise to the data according to the forward diffusion process.}
        \label{fig:trajectory_densities:diffusion}
    \end{subfigure}
    \par\medskip
    \begin{subfigure}{\linewidth}
        \begin{center}
            \TrajectoryDensityTikz{cfg_0} \hfill \TrajectoryDensityTikz{cfg_1} \hfill \TrajectoryDensityTikz{cfg_2}
        \end{center}
        \caption{\ac{cfg}-guided trajectory densities obtained by running the diffusion reverse process with \acl{cfg}.}
        \label{fig:trajectory_densities:guidance}
    \end{subfigure}
    \caption{One-dimensional illustration of the mismatch between standard diffusion trajectories used for training in existing guidance distillation methods and the actual guided trajectories followed during inference. The \ac{cfg} trajectories in (c) occupy regions in space distinct from the standard diffusion trajectories in (b). Training directly on \ac{cfg}-guided trajectories ensures that the adapters focus on the regions primarily encountered during sampling with \ac{cfg}.}
    \label{fig:trajectory_densities}
\end{figure}%
}

\newcommand{\AdapterAndBaseModel}{%
\begin{figure}[t]
    \setlength{\twemojiDefaultHeight}{1.2em}
    \centering
    \begin{tikzpicture}[
          box_adapter/.style={rectangle, draw, fill=Blue, node distance=1cm, minimum height=1.5em, minimum width=1.5em, text centered},
          box_attach/.style={rectangle, fill=Cyan, draw, node distance=1cm, minimum width=1.5em, minimum height=1.5em, text centered},
          box_block/.style={rectangle, fill=Green, draw, node distance=1cm, text width=8em, minimum height=1.5em, text centered},
          box_head/.style={rectangle, fill=Yellow, draw, node distance=1cm, text width=8em, minimum height=1.5em, text centered},
          line/.style={draw, thick},
          arrow/.style={draw, thick, ->},
    ]
        \node at (0, 0) (input) {$\vec{x}_t, t, c$};

        \node [box_block, above=0.5 of input] (block1) {block};
        \node [box_attach, above=0.1 of block1] (ca1) {+};
        \node [box_adapter, left=0.6 of block1] (adapter1) {adapter};
        \node [below=0.1 of block1, inner sep=0pt, outer sep=0pt] (adapter_input1) {};
        
        \path [line] (block1) -- (ca1);
        \path [arrow] (adapter_input1) -| (adapter1) |- (ca1);

        \node [box_block, above=0.5 of ca1] (block2) {block};
        \node [box_attach, above=0.1 of block2] (ca2) {+};
        \node [box_adapter, left=0.6 of block2] (adapter2) {adapter};
        \node [below=0.1 of block2, inner sep=0pt, outer sep=0pt] (adapter_input2) {};
        
        \path [line] (block2) -- (ca2);
        \path [arrow] (adapter_input2) -| (adapter2) |- (ca2);

        \node [box_block, above=1.15 of ca2] (block3) {block};
        \node [box_attach, above=0.1 of block3] (ca3) {+};
        \node [box_adapter, left=0.6 of block3] (adapter3) {adapter};
        \node [below=0.1 of block3, inner sep=0pt, outer sep=0pt] (adapter_input3) {};

        \node [above=0.2 of ca2, inner sep=0pt] (ca2_out) {};
        \path [line] (ca2) -- (ca2_out);
        \node [below=0.2 of block3, inner sep=0pt] (block3_in) {};
        \path [line] (block3_in) -- (block3);
        
        \path [line] (block3) -- (ca3);
        \path [arrow] (adapter_input3) -| (adapter3) |- (ca3);

        \node [box_head, above=0.5 of ca3] (head) {head};
        \node [above=0.5 of head] (output) {$\tilde{\vec{\epsilon}}^{\omega}_{[\vec{\theta},\vec{\psi}]}$};

        \node[left=3.0 of input] (guidance scale) {$\omega$};

        \path [line] (input) -- (block1);
        \path [line] (ca1) -- (block2);
        \path [line, dotted] (ca2_out) -- (block3_in);
        \path [line] (ca3) -- (head);
        \path [arrow] (head) -- (output);

        \path [line] (guidance scale) |- (adapter1);
        \path [line] (guidance scale) |- (adapter2);
        \path [line] (guidance scale) |- (adapter3);

        \node [draw, dashed, inner sep=2mm, fit=(adapter1) (adapter2) (adapter3)] (adapter_box) {};
        \node [above=0.3 of adapter_box.north west] (adapter_params) {\footnotesize \textit{Trainable $\vec{\psi}$}};
        \path [draw] (adapter_box) -- (adapter_params);
        \node [above=0 of adapter_params] {\twemoji{fire}};
        
        \node [draw, dashed, inner sep=2mm, fit=(block1) (ca1) (block2) (ca2) (block3) (ca3) (head)] (model_box) {};
        \node [right=0 of model_box] (base_params) {\footnotesize \textit{Frozen $\vec{\theta}$}};
        \node [above=0 of base_params] {\twemoji{snowflake}};
    \end{tikzpicture}
    \caption{Visual illustration of the trainable adapters alongside the frozen base model. The adapters are typically integrated with attention layers (either self-attention or cross-attention), and their outputs are added to those of the frozen attention blocks.}
    \label{fig:adapter}
\end{figure}%
}

\newcommand{\rulesep}{\unskip\ \vrule\ }

\newcommand{\TrainingInference}{%
\begin{figure*}[t]
    \centering
    \begin{subfigure}[t]{0.3\linewidth}
        \centering
        \begin{tikzpicture}[
            line/.style={draw, thick},
            arrow/.style={draw, thick, ->},
        ]
            \node [inner sep=0pt, outer sep=0pt] at (0, 0) (origin) {};

            \begin{scope}[zlevel=main]
                \node [right=1 of origin, draw, minimum width=6em, minimum height=4em, fill=Green] (model) {Base model};
            \end{scope}
            
            \node [above=0.1 of origin] (input1) {$\vec{x}_t, t, c$};
            \node [below=0.1 of origin] (input2) {$\vec{x}_t, t, \emptyset$};

            \node [circle, minimum width=1.2em, draw, fill=Blue, right=0.2 of model, inner sep=0pt, outer sep=0pt] (output) {\small $\omega$};
            \node [right=0.25 of output] (output_var) {$\tilde{\vec{\epsilon}}$};

            \node [above=0.3 of model] (next_step) {$\vec{x}_{t-1}$};

            \begin{scope}[zlevel=-1]
                \path [line] (input1) -| (output);
                \path [line] (input2) -| (output);
                \path [arrow] (output) -- (output_var);
                \path [draw, dashed] (output_var) |- (next_step);
                \path [draw, ->, dashed] (next_step) -| (input1);
            \end{scope}
        \end{tikzpicture}
        \par\bigskip
        \caption{Trajectory collection (\Cref{sec:trajectories}).}
        \label{fig:trajectory_collection}
    \end{subfigure}
    \rulesep
    \begin{subfigure}[t]{0.36\linewidth}
        \centering
        \begin{tikzpicture}[
            line/.style={draw, thick},
            arrow/.style={draw, thick, ->},
        ]
            \node at (0, 0) (input) {$\vec{x}_t, t, c, \omega$};
            
            \begin{scope}[zlevel=main]
                \node [right=0.35 of input, draw, minimum width=6em, minimum height=4em, fill=Cyan] (model) {Base + adapter};
            \end{scope}

            \node [right=0.35 of model] (output) {$\hat{\vec{\epsilon}}$};
            \node [right=0.3 of output] (loss) {$\mathcal{L}$};
            \node [above=0.3 of loss] (y) {$\tilde{\vec{\epsilon}}$};

            \begin{scope}[zlevel=-1]
                \path [line] (input) -- (model);
                \path [arrow] (model) -- (output);
                \path [arrow] (output) -- (loss);
                \path [arrow] (y) -- (loss);
                \path [draw, dashed, ->] (loss) --++(0.4,0) --++(0,1.3) -| (model);
            \end{scope}
        \end{tikzpicture}
        \par\bigskip
        \caption{Training on collected trajectories (\Cref{sec:adapters}).}
        \label{fig:training}
    \end{subfigure}
    \rulesep
    \begin{subfigure}[t]{0.3\linewidth}
        \centering
        \begin{tikzpicture}[
            line/.style={draw, thick},
            arrow/.style={draw, thick, ->},
        ]
            \node at (0, 0) (input) {$\vec{x}_t, t, c, \omega$};
            
            \begin{scope}[zlevel=main]
                \node [right=0.35 of input, draw, minimum width=6em, minimum height=4em, fill=Cyan] (model) {Base + adapter};
            \end{scope}

            \node [right=0.35 of model] (output) {$\tilde{\vec{\epsilon}}$};
            \node [above=0.3 of model] (next_step) {$\vec{x}_{t-1}$};

            \begin{scope}[zlevel=-1]
                \path [line] (input) -- (model);
                \path [arrow] (model) -- (output);
                \path [draw, dashed] (output) |- (next_step);
                \path [draw, ->, dashed] (next_step) -| (input);
            \end{scope}
        \end{tikzpicture}
        \par\bigskip
        \caption{Inference with adapters.}
        \label{fig:inference}
    \end{subfigure}
    \caption{High-level overview of AGD components. (a) Instead of training on diffusion trajectories, we first run the sampling process with classifier-free guidance (CFG) and use the resulting guided trajectories (\ie, intermediate predictions at each time step $t$) as our training dataset. (b) We then introduce small adapters to the base model and train them to replicate the CFG-guided predictions from (a) while keeping the base model frozen. (c) During inference, the base model combined with the trained adapter produces guided predictions in a single forward pass, effectively doubling the sampling speed compared to CFG.}
    \label{fig:trajectory_training_inference}
\end{figure*}
}

\newcommand{\GuidanceVsFID}{%
\begin{figure*}[ht]
    \centering
    \begin{tikzpicture}
        \begin{axis}[width=0.28\linewidth, height=0.18\linewidth, enlarge x limits=0, xlabel={$\omega$}, title={FID}, scale only axis, ymajorgrids=true,
            xmajorgrids=true,
            grid style=dashed,
            major grid style={lightgray}]
            \addplot [
                C3,
                thick,
                mark=none,
            ] coordinates { (1.0, 12.5714) (1.1, 9.1329) (1.2, 6.9048) (1.3, 5.8001) (1.4, 5.2191) (1.5, 5.3029) (1.6, 5.5621) (1.7, 6.3511) (1.8, 7.3443) (1.9, 8.0032) (2.0, 8.8390) (2.1, 9.5655) (2.2, 10.5575) (2.3, 11.2183) (2.4, 12.3712) (2.5, 13.5309) (2.6, 13.9942) (2.7, 14.6184) (2.8, 15.5816) (2.9, 15.5114) (3.0, 16.5300) };
            \addplot [
                C0,
                thick,
                mark=none,
            ] coordinates { (1.0, 11.7539) (1.1, 10.6220) (1.2, 9.4493) (1.3, 7.9503) (1.4, 7.1325) (1.5, 6.3838) (1.6, 5.8381) (1.7, 5.4676) (1.8, 5.3233) (1.9, 5.1101) (2.0, 5.0335) (2.1, 5.1267) (2.2, 5.3288) (2.3, 5.2604) (2.4, 5.2440) (2.5, 5.5853) (2.6, 5.7375) (2.7, 5.8385) (2.8, 6.2252) (2.9, 6.0819) (3.0, 6.3916) };
        \end{axis}
    \end{tikzpicture}
    \hfill
    \begin{tikzpicture}
        \begin{axis}[width=0.28\linewidth, height=0.18\linewidth, enlarge x limits=0, xlabel={$\omega$}, title={Precision}, scale only axis, ymajorgrids=true,
            xmajorgrids=true,
            grid style=dashed,
            major grid style = {lightgray},]
            \addplot [
                C3,
                thick,
                mark=none,
            ] coordinates { (1.0, 0.6681) (1.1, 0.7120) (1.2, 0.7485) (1.3, 0.7749) (1.4, 0.8040) (1.5, 0.8325) (1.6, 0.8458) (1.7, 0.8627) (1.8, 0.8721) (1.9, 0.8871) (2.0, 0.8970) (2.1, 0.8998) (2.2, 0.9059) (2.3, 0.9098) (2.4, 0.9191) (2.5, 0.9224) (2.6, 0.9232) (2.7, 0.9234) (2.8, 0.9256) (2.9, 0.9266) (3.0, 0.9314) };
            \addplot [
                C0,
                thick,
                mark=none,
            ] coordinates { (1.0, 0.6772) (1.1, 0.6892) (1.2, 0.7091) (1.3, 0.7226) (1.4, 0.7434) (1.5, 0.7612) (1.6, 0.7664) (1.7, 0.7895) (1.8, 0.7902) (1.9, 0.8041) (2.0, 0.7990) (2.1, 0.8070) (2.2, 0.8199) (2.3, 0.8222) (2.4, 0.8349) (2.5, 0.8340) (2.6, 0.8398) (2.7, 0.8420) (2.8, 0.8455) (2.9, 0.8463) (3.0, 0.8509) };
        \end{axis}
    \end{tikzpicture}
    \hfill
    \begin{tikzpicture}
        \begin{axis}[width=0.28\linewidth, height=0.18\linewidth, enlarge x limits=0, xlabel={$\omega$}, title={Recall}, scale only axis, ymajorgrids=true,
            xmajorgrids=true,
            grid style=dashed,
            major grid style = {lightgray},]
            \addplot [
                C3,
                thick,
                mark=none,
            ] coordinates { (1.0, 0.7435) (1.1, 0.7318) (1.2, 0.7171) (1.3, 0.6977) (1.4, 0.6765) (1.5, 0.6583) (1.6, 0.6501) (1.7, 0.6268) (1.8, 0.5941) (1.9, 0.5822) (2.0, 0.5583) (2.1, 0.5524) (2.2, 0.5311) (2.3, 0.5158) (2.4, 0.4850) (2.5, 0.4845) (2.6, 0.4638) (2.7, 0.4528) (2.8, 0.4384) (2.9, 0.4310) (3.0, 0.4249) };

            \addplot [
                C0,
                thick,
                mark=none,
            ] coordinates { (1.0, 0.7421) (1.1, 0.7393) (1.2, 0.7341) (1.3, 0.7262) (1.4, 0.7175) (1.5, 0.7014) (1.6, 0.6981) (1.7, 0.6997) (1.8, 0.6847) (1.9, 0.6763) (2.0, 0.6778) (2.1, 0.6674) (2.2, 0.6693) (2.3, 0.6591) (2.4, 0.6542) (2.5, 0.6434) (2.6, 0.6388) (2.7, 0.6351) (2.8, 0.6360) (2.9, 0.6243) (3.0, 0.6236) };
        \end{axis}
    \end{tikzpicture}
    \par\smallskip
    \begin{tikzpicture}
        \footnotesize
        \begin{axis}[hide axis, width=\linewidth, legend columns=11, xmin=0, xmax=1, ymin=0, ymax=1, legend style={at={(0.0, 0.0)}, anchor=north}]
            \addplot [C3,thick, mark=*, mark size={2pt}] coordinates {(0, -1)};
            \addplot [C0, thick, mark=*, mark size={2pt}] coordinates {(0, -1)};
            \legend{CFG,AGD (Ours)};
        \end{axis}
    \end{tikzpicture}
    \caption{The performance of \ac{dit} with \ac{agd} as we increase the guidance scale. Compared to \ac{cfg}, \ac{agd} offers a better trade-off between precision and recall resulting in better \ac{fid} for most guidance scales.}
    \label{fig:guidance_vs_fid}
\end{figure*}%
}

\newcommand{\Samples}{
\begin{figure*}[ht]
    \centering
    \begin{subfigure}[t]{\linewidth}
        \centering
        \begin{tikzpicture}[every node/.style={inner sep=0pt, outer sep=0pt}]
            \node (img1) {\includegraphics[width=0.097\textwidth]{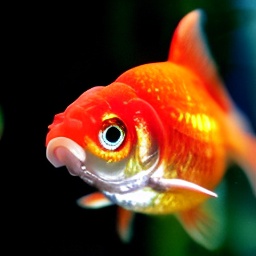}};
            \node [right=0 of img1] (img2) {\includegraphics[width=0.097\textwidth]{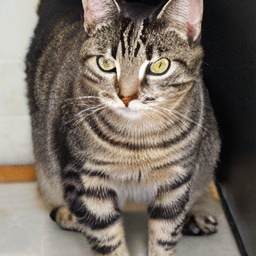}};
            \node [right=0 of img2] (img3) {\includegraphics[width=0.097\textwidth]{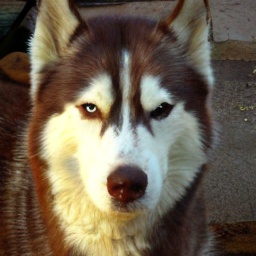}};
            \node [right=0 of img3] (img4) {\includegraphics[width=0.097\textwidth]{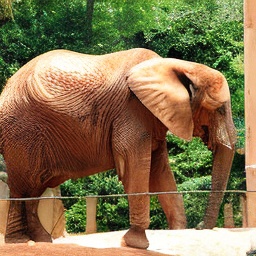}};
            \node [right=0 of img4] (img5) {\includegraphics[width=0.097\textwidth]{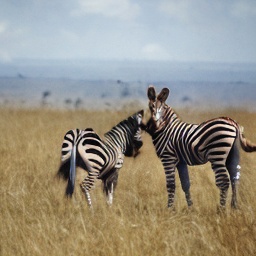}};
            \node [below=0 of img1] (img6) {\includegraphics[width=0.097\textwidth]{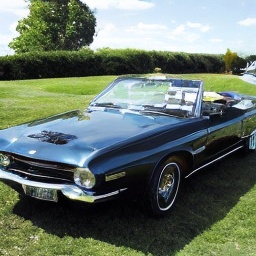}};
            \node [right=0 of img6] (img7) {\includegraphics[width=0.097\textwidth]{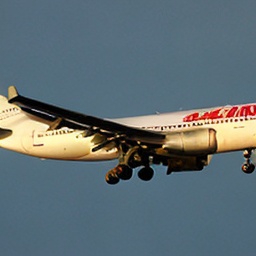}};
            \node [right=0 of img7] (img8) {\includegraphics[width=0.097\textwidth]{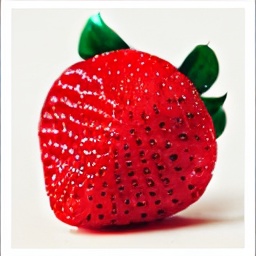}};
            \node [right=0 of img8] (img9) {\includegraphics[width=0.097\textwidth]{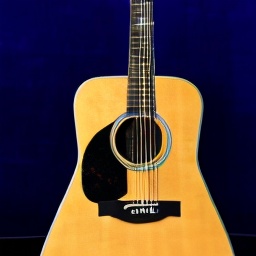}};
            \node [right=0 of img9] (img10) {\includegraphics[width=0.097\textwidth]{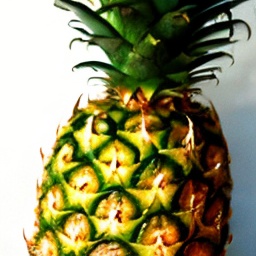}};
            
            \node [right=0.5 of img5] (img11) {\includegraphics[width=0.097\textwidth]{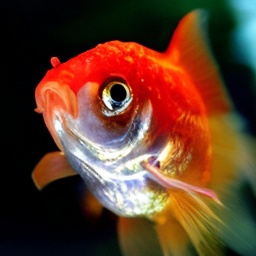}};
            \node [right=0 of img11] (img12) {\includegraphics[width=0.097\textwidth]{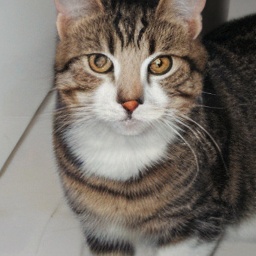}};
            \node [right=0 of img12] (img13) {\includegraphics[width=0.097\textwidth]{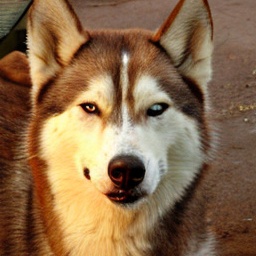}};
            \node [right=0 of img13] (img14) {\includegraphics[width=0.097\textwidth]{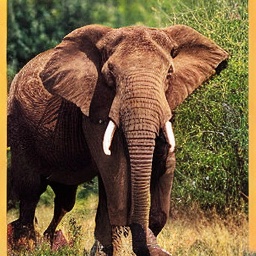}};
            \node [right=0 of img14] (img15) {\includegraphics[width=0.097\textwidth]{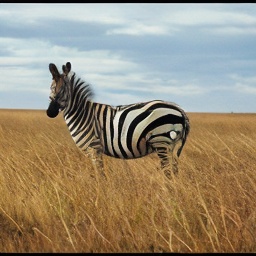}};
            \node [below=0 of img11] (img16) {\includegraphics[width=0.097\textwidth]{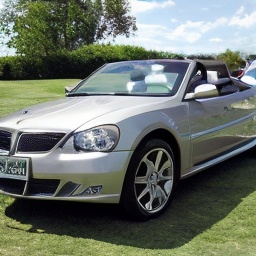}};
            \node [right=0 of img16] (img17) {\includegraphics[width=0.097\textwidth]{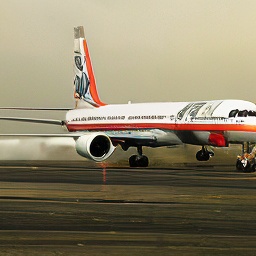}};
            \node [right=0 of img17] (img18) {\includegraphics[width=0.097\textwidth]{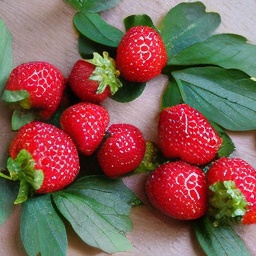}};
            \node [right=0 of img18] (img19) {\includegraphics[width=0.097\textwidth]{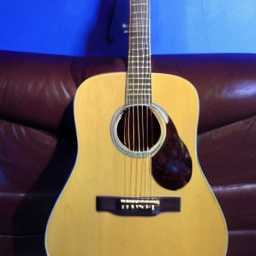}};
            \node [right=0 of img19] (img20) {\includegraphics[width=0.097\textwidth]{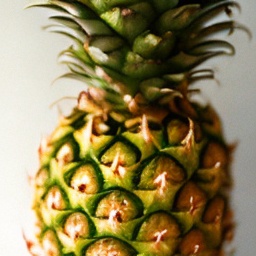}};
            
            \node [fit=(img1) (img2) (img3) (img4) (img5) (img6) (img7) (img8) (img9) (img10)] (cfg_box) {};
            \node [above=0.15 of cfg_box] {\acs{cfg}};
            
            \node [fit=(img11) (img12) (img13) (img14) (img15) (img16) (img17) (img18) (img19) (img20)] (agd_box) {};
            \node [above=0.15 of agd_box] {\acs{agd} (Ours)};
        \end{tikzpicture}
        \caption{\Acl{dit} (256$\times$256)}
        \label{fig:samples:dit}
    \end{subfigure}
    \par\medskip
    \begin{subfigure}[t]{\linewidth}
        \centering
        \begin{tikzpicture}[every node/.style={inner sep=0pt, outer sep=0pt}]
            \node (img1) {\includegraphics[width=0.194\textwidth]{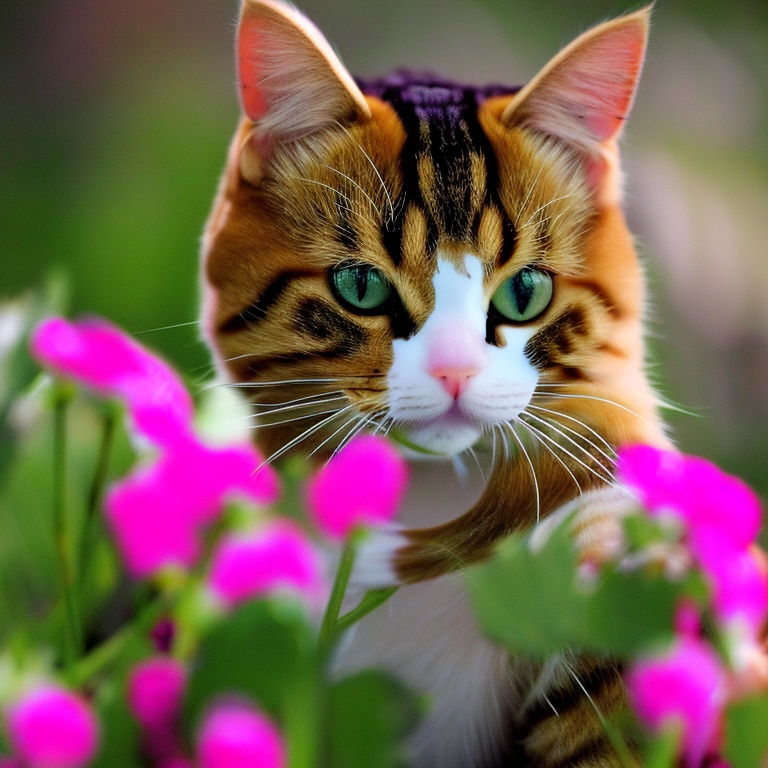}};
            \node [right=0 of img1] (img2) {\includegraphics[width=0.194\textwidth]{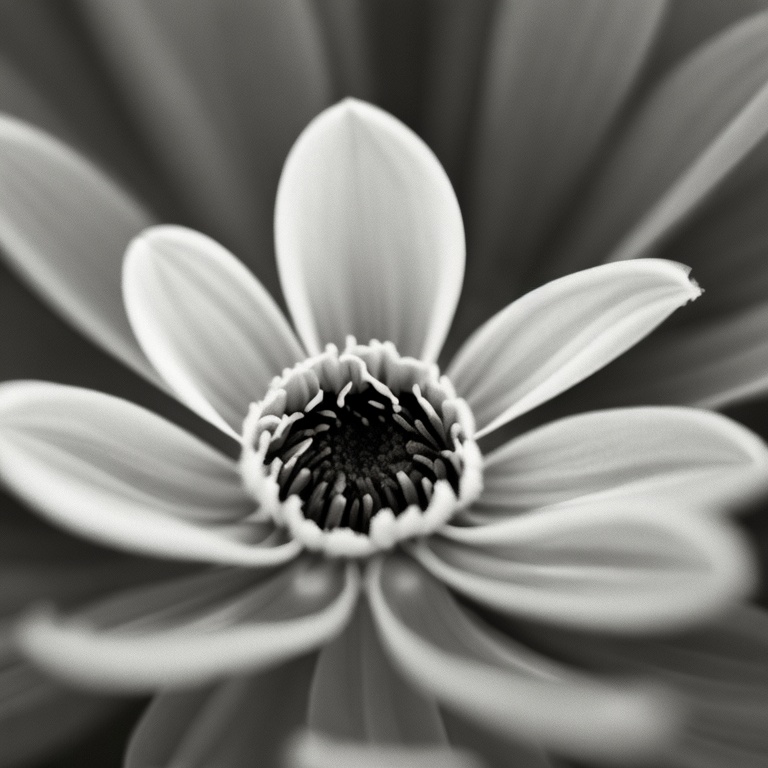}};
            \node [right=0 of img2.north east, anchor=north west] (img3) {\includegraphics[width=0.097\textwidth]{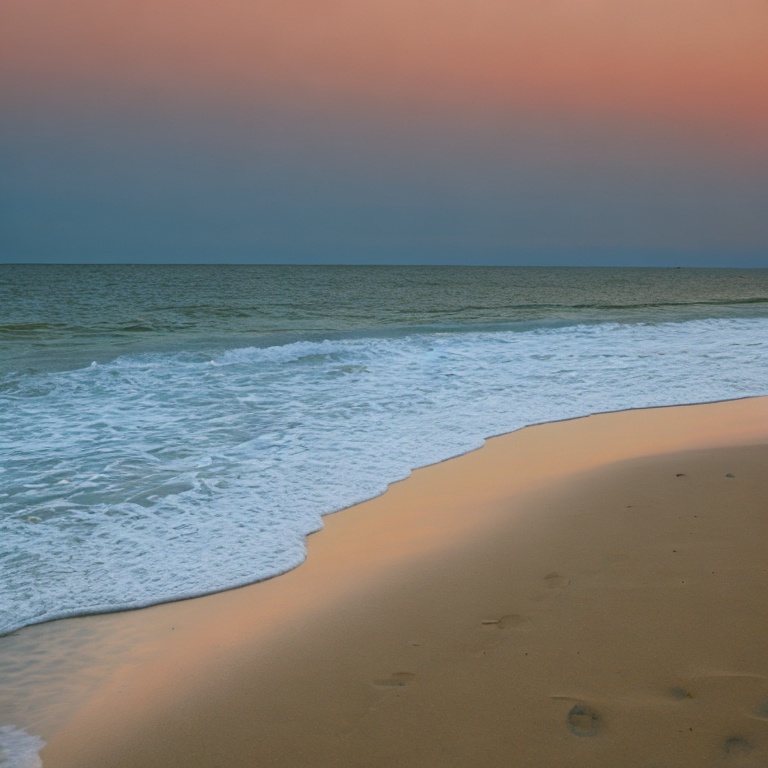}};
            \node [below=0 of img3] (img4) {\includegraphics[width=0.097\textwidth]{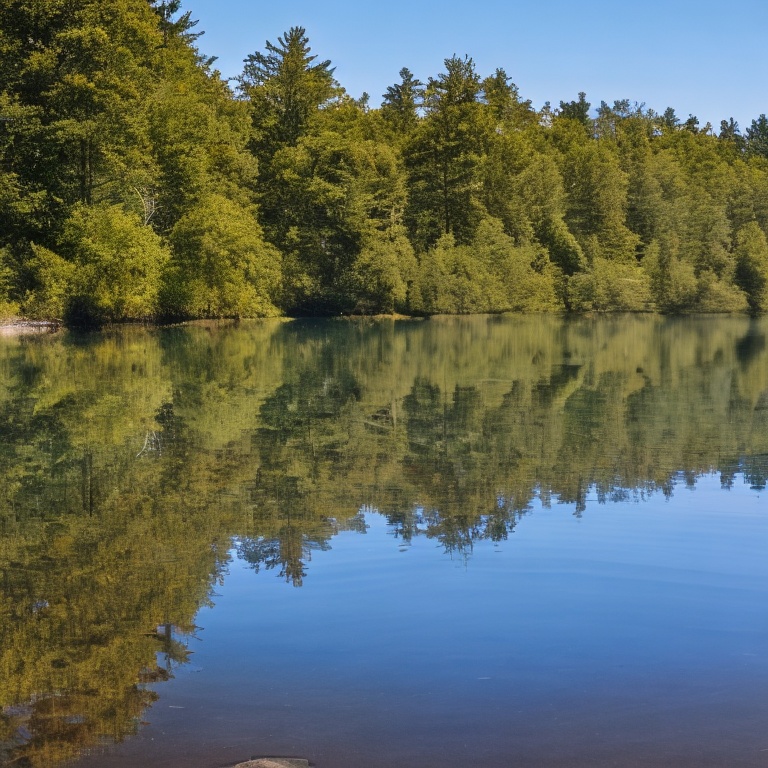}};
            
            \node [right=0.5 of img3.north east, anchor=north west] (img5) {\includegraphics[width=0.194\textwidth]{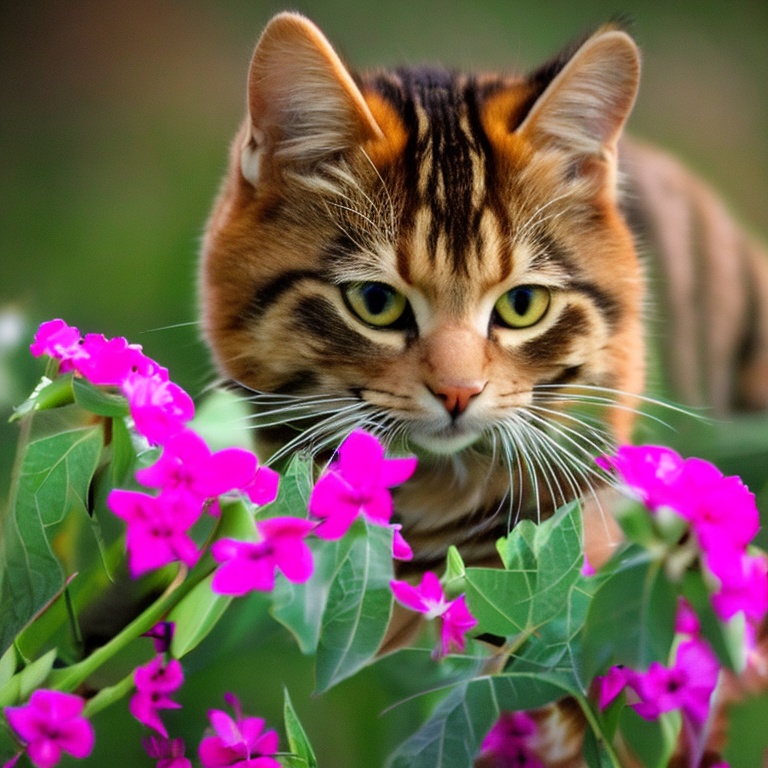}};
            \node [right=0 of img5] (img6) {\includegraphics[width=0.194\textwidth]{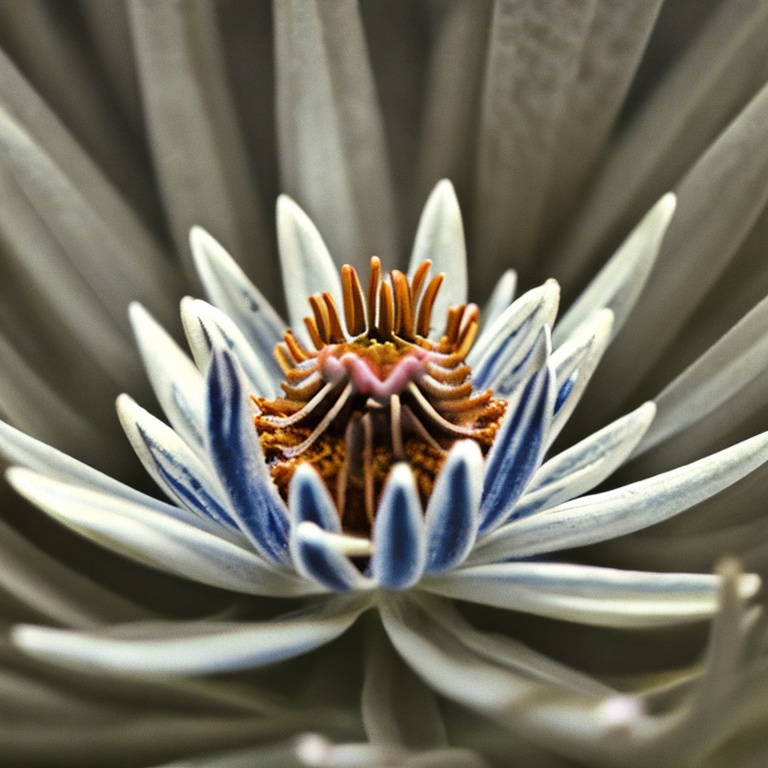}};
            \node [right=0 of img6.north east, anchor=north west] (img7) {\includegraphics[width=0.097\textwidth]{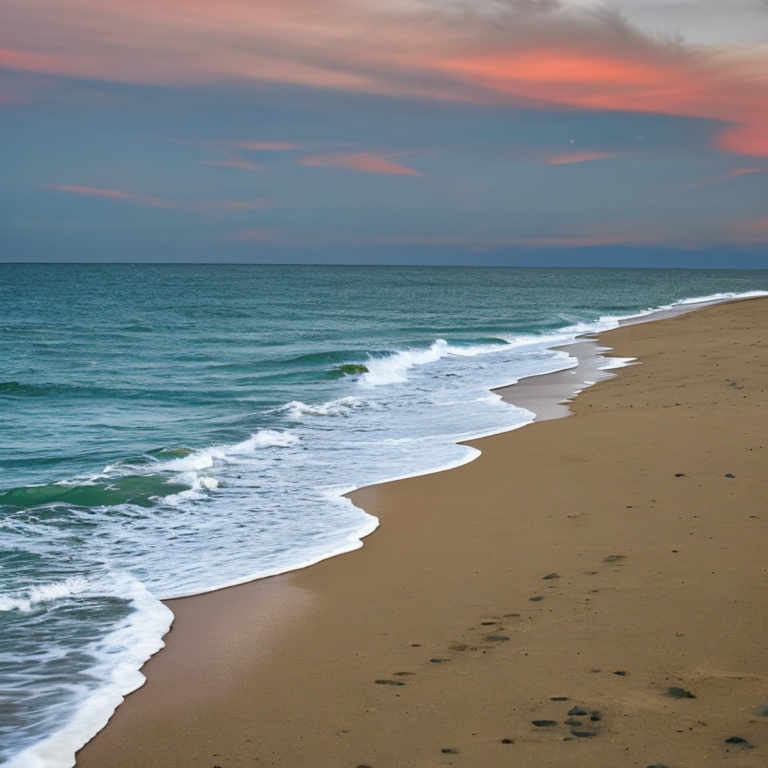}};
            \node [below=0 of img7] (img8) {\includegraphics[width=0.097\textwidth]{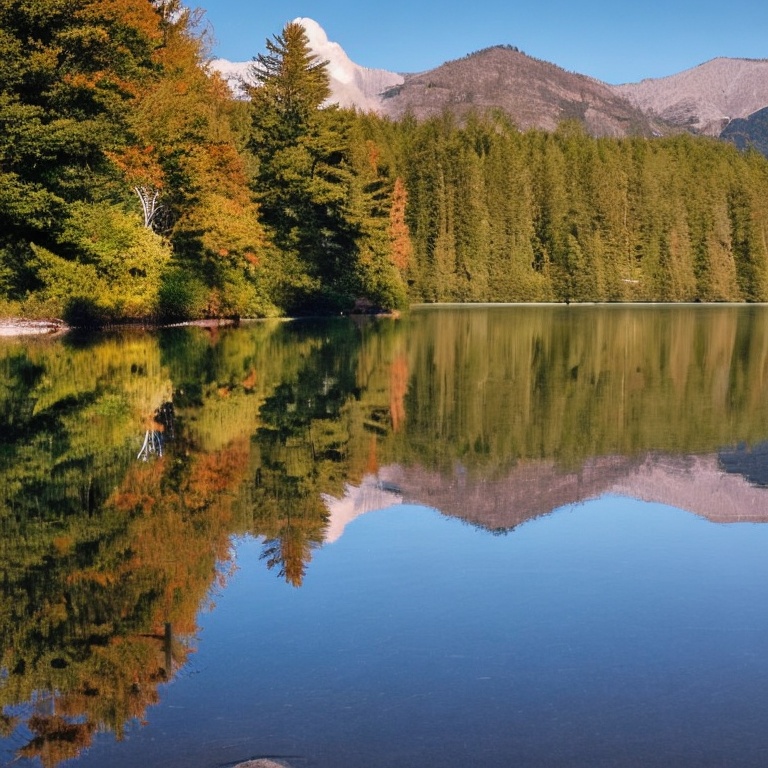}};
            
            \node [fit=(img1) (img2) (img3) (img4)] (cfg_box) {};
            \node [above=0.15 of cfg_box] {\acs{cfg}};
            
            \node [fit=(img5) (img6) (img7) (img8)] (agd_box) {};
            \node [above=0.15 of agd_box] {\acs{agd} (Ours)};
        \end{tikzpicture}
        \caption{\Acl{sd2} (768$\times$768)}
        \label{fig:samples:sd2}
    \end{subfigure}
    \par\medskip
    \begin{subfigure}[t]{\linewidth}
        \centering
        \begin{tikzpicture}[every node/.style={inner sep=0pt, outer sep=0pt}]
            \node (img1) {\includegraphics[width=0.194\textwidth]{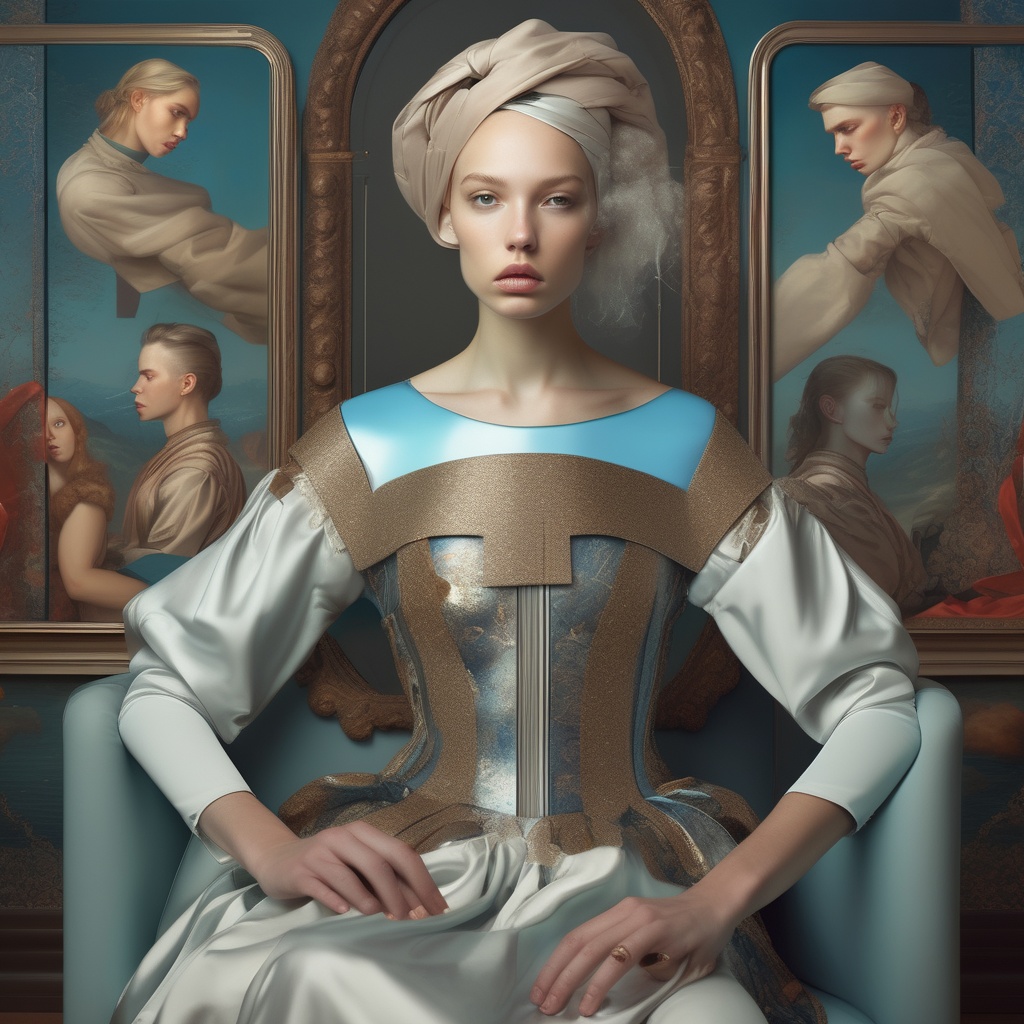}};
            \node [right=0 of img1] (img2) {\includegraphics[width=0.194\textwidth]{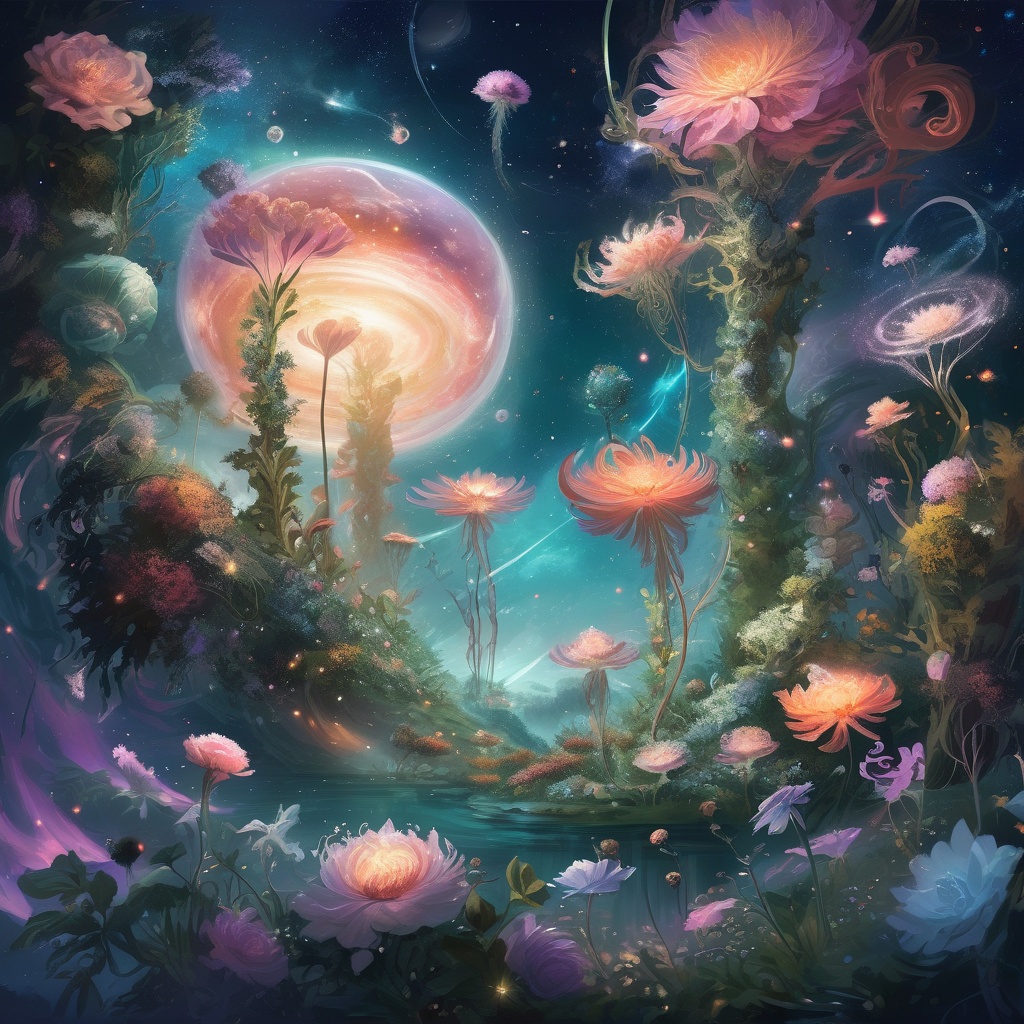}};
            \node [right=0 of img2.north east, anchor=north west] (img3) {\includegraphics[width=0.097\textwidth]{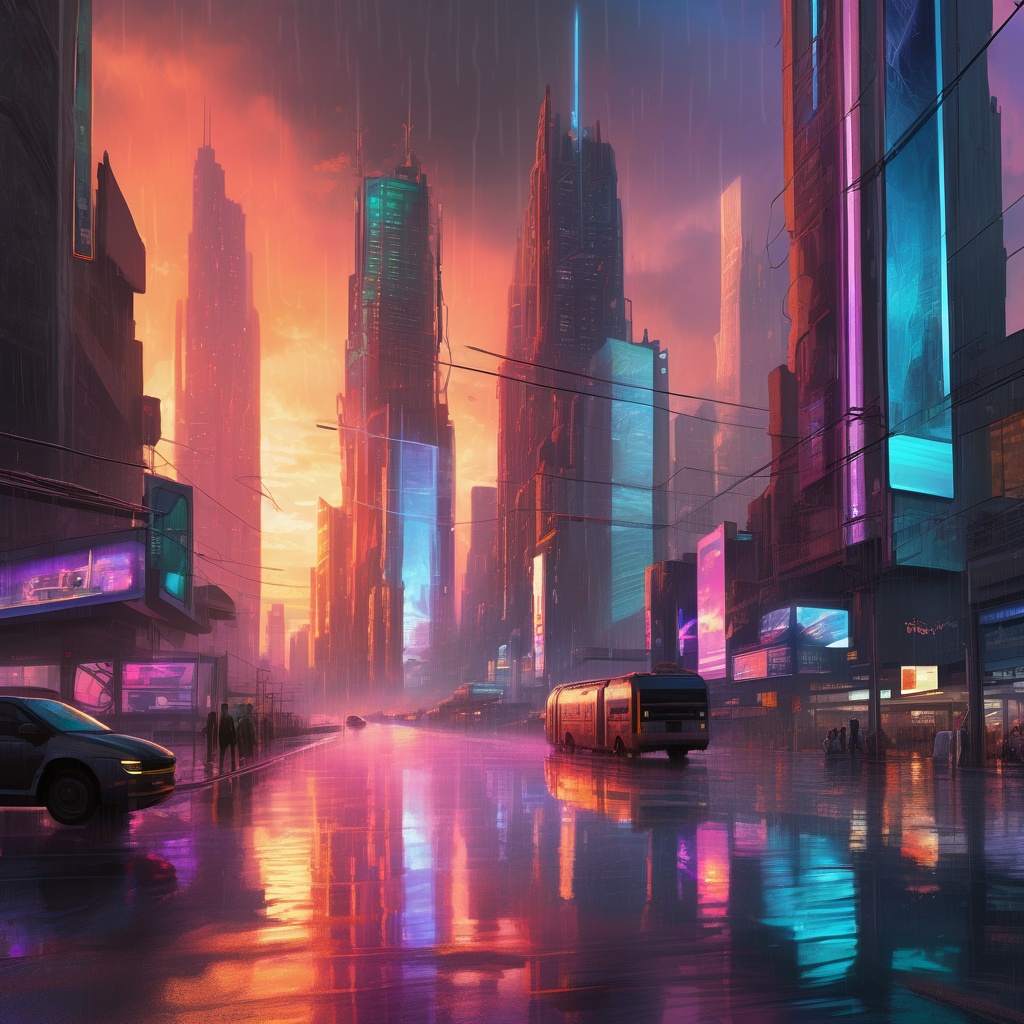}};
            \node [below=0 of img3] (img4) {\includegraphics[width=0.097\textwidth]{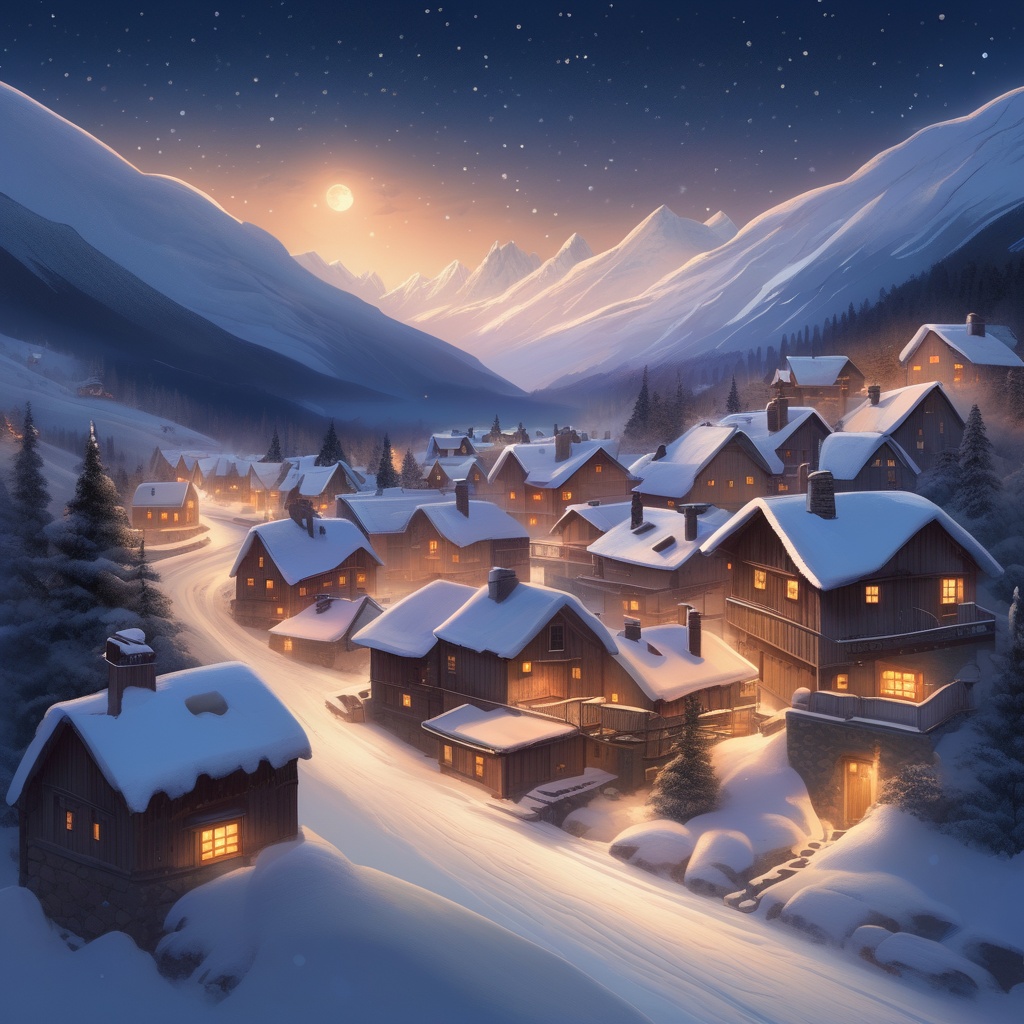}};
            
            \node [right=0.5 of img3.north east, anchor=north west] (img5) {\includegraphics[width=0.194\textwidth]{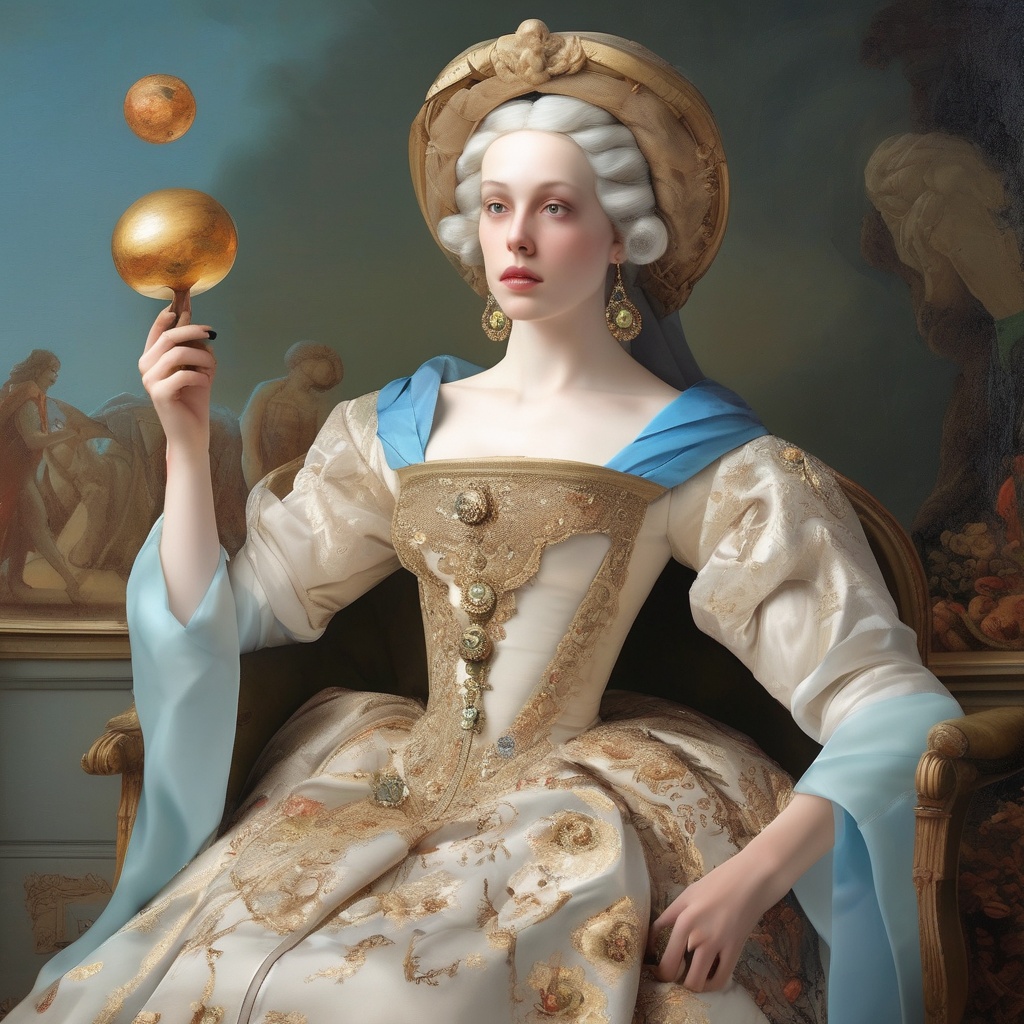}};
            \node [right=0 of img5] (img6) {\includegraphics[width=0.194\textwidth]{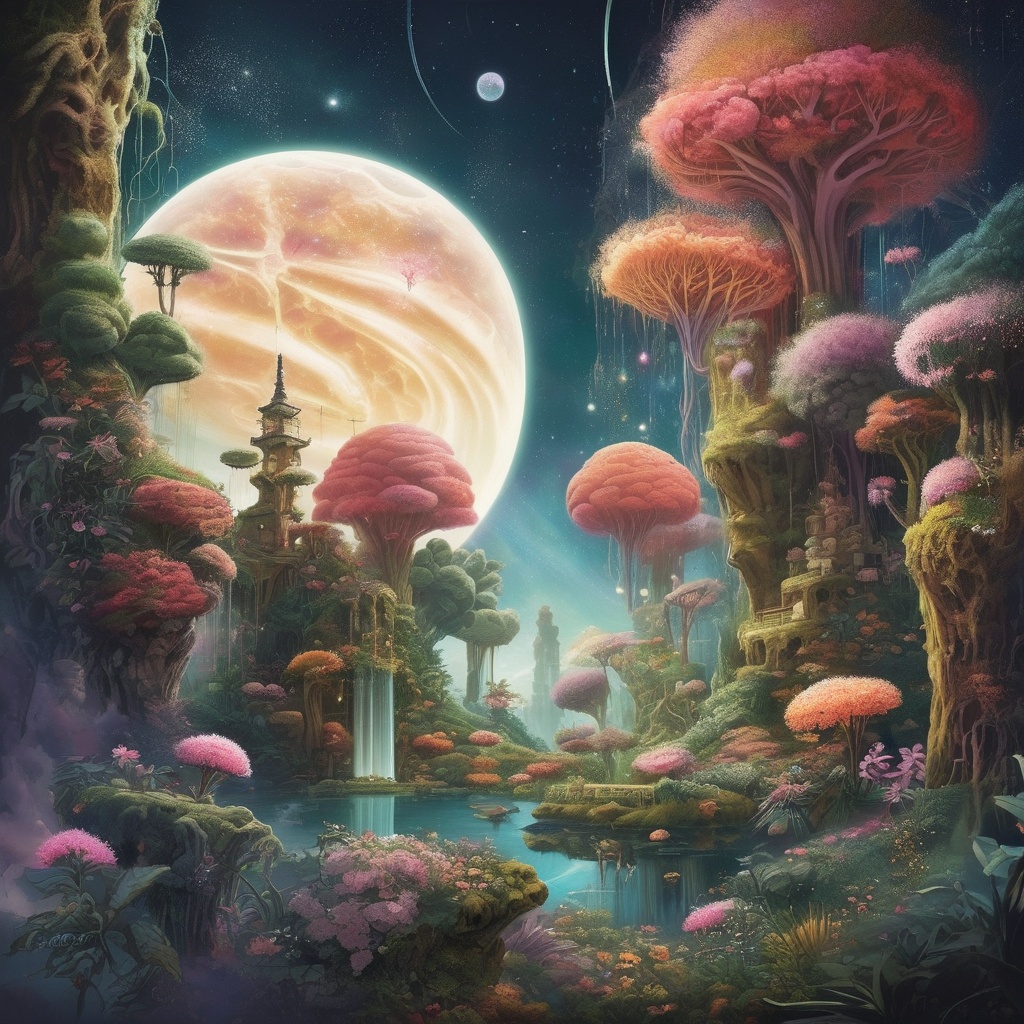}};
            \node [right=0 of img6.north east, anchor=north west] (img7) {\includegraphics[width=0.097\textwidth]{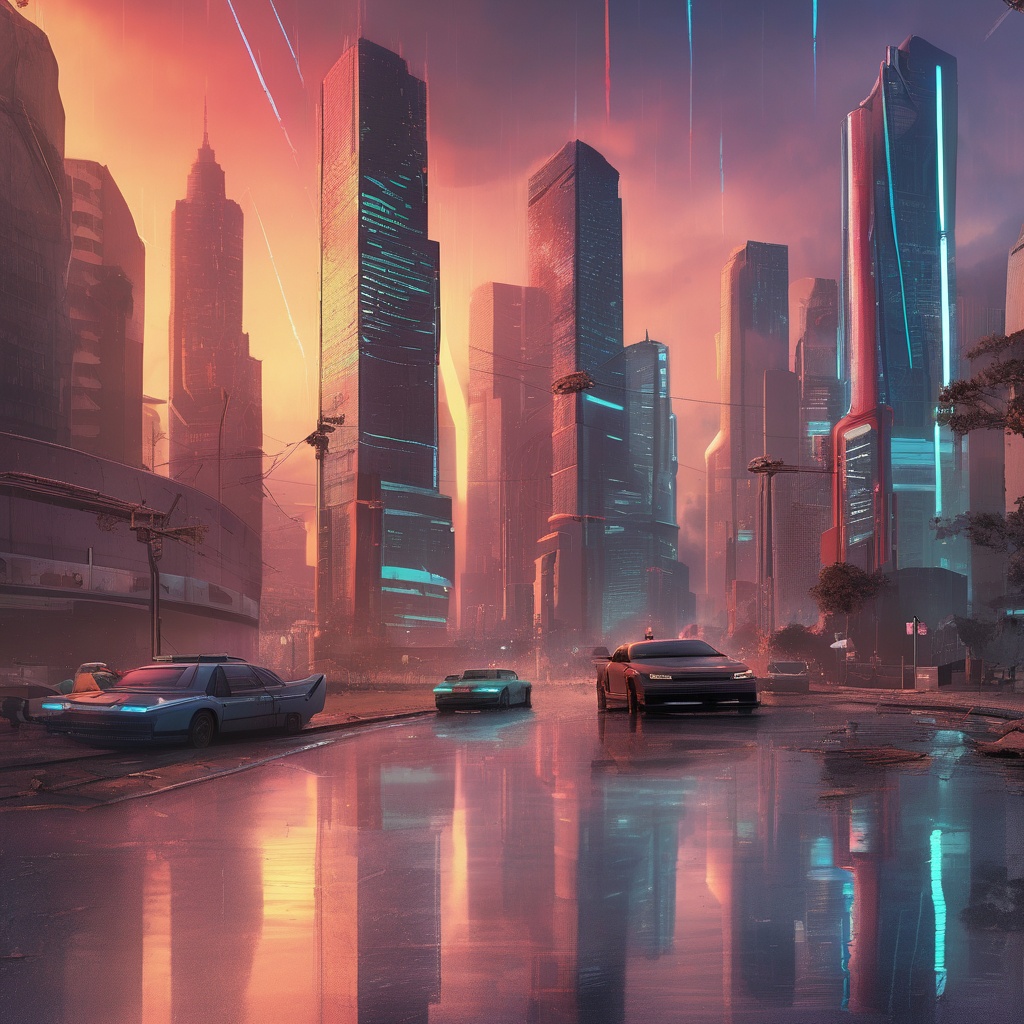}};
            \node [below=0 of img7] (img8) {\includegraphics[width=0.097\textwidth]{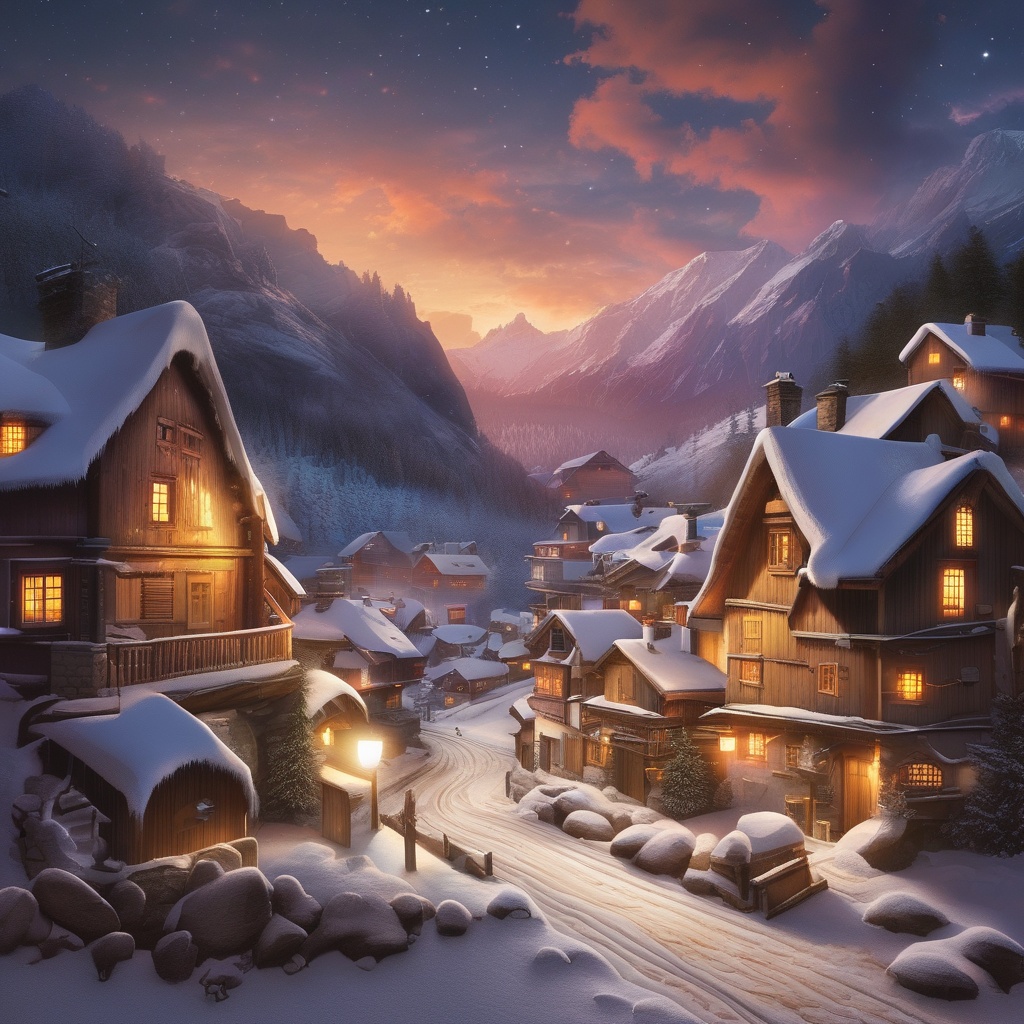}};

            \node [fit=(img1) (img2) (img3) (img4)] (cfg_box) {};
            \node [above=0.15 of cfg_box] {\acs{cfg}};
            
            \node [fit=(img5) (img6) (img7) (img8)] (agd_box) {};
            \node [above=0.15 of agd_box] {\acs{agd} (Ours)};
        \end{tikzpicture}
        \caption{\Acl{sdxl} (1024$\times$1024)}
        \label{fig:samples:sdxl}
    \end{subfigure}
    \caption{Qualitative comparison between \ac{agd} and \ac{cfg}. \Ac{agd} produces samples with comparable quality to \ac{cfg} while achieving twice the inference speed by requiring only a single forward pass through the model. Additionally, \ac{agd} samples maintain structural similarity to \ac{cfg} but often have better visual coherence.}
    \label{fig:samples}
\end{figure*}
}

\newcommand{\SamplesOOD}{
    \begin{figure}[t]
        \centering
        \begin{tikzpicture}[every node/.style={inner sep=0pt, outer sep=0pt}]
            \node (img1) {\includegraphics[width=0.23\linewidth]{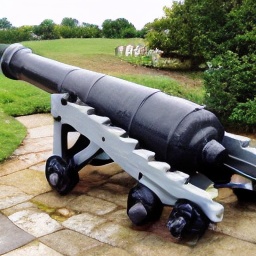}};
            \node [right=0 of img1] (img2) {\includegraphics[width=0.23\linewidth]{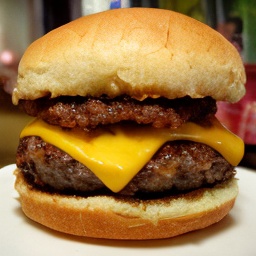}};
            \node [below=0 of img1] (img3) {\includegraphics[width=0.23\linewidth]{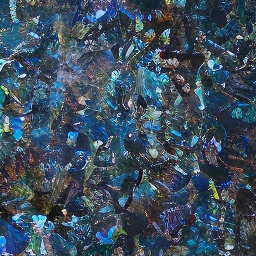}};
            \node [right=0 of img3] (img4) {\includegraphics[width=0.23\linewidth]{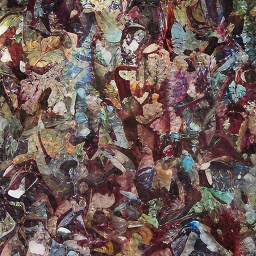}};
            
            \node [right=0.2 of img2] (img11) {\includegraphics[width=0.23\linewidth]{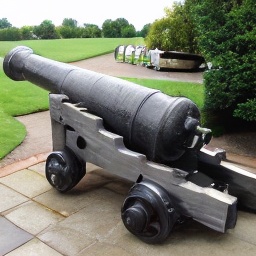}};
            \node [right=0 of img11] (img12) {\includegraphics[width=0.23\linewidth]{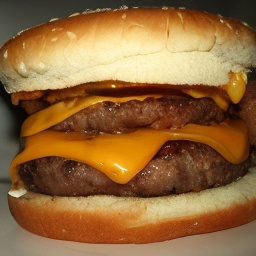}};
            \node [below=0 of img11] (img13) {\includegraphics[width=0.23\linewidth]{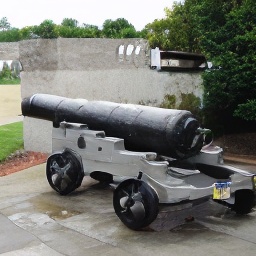}};
            \node [right=0 of img13] (img14) {\includegraphics[width=0.23\linewidth]{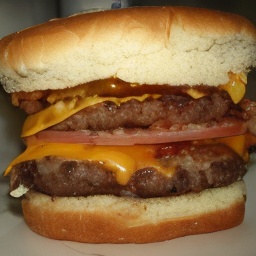}};
            
            \node [rotate=90, left=0.3 of img1, anchor=center] {$\omega=4$};
            \node [rotate=90, left=0.3 of img3, anchor=center]  {$\omega=10$};
            
            \node [fit=(img1) (img2) (img3) (img4)] (gd_box) {};
            \node [above=0.15 of gd_box] {GD \citep{meng2023distillation}};
            
            \node [fit=(img11) (img12) (img13) (img14)] (agd_box) {};
            \node [above=0.15 of agd_box] {AGD (Ours)};
        \end{tikzpicture}
        \caption{Comparison of \ac{agd} with \acf{gd} \citep{meng2023distillation} for unseen guidance scales. While \ac{gd} fails completely for out-of-domain guidance scales, \ac{agd} continues to generate meaningful images. The models in this experiment were trained for $\wcfg \in [1, 6]$.}
        \label{fig:ood}
    \end{figure}
}

\newcommand{\SamplesScheduler}{
    \begin{figure}[t]
        \centering
        \vspace{-0.3cm}
        \begin{tikzpicture}[every node/.style={inner sep=0pt, outer sep=0pt}]
            \node (img1) {\includegraphics[width=0.23\linewidth]{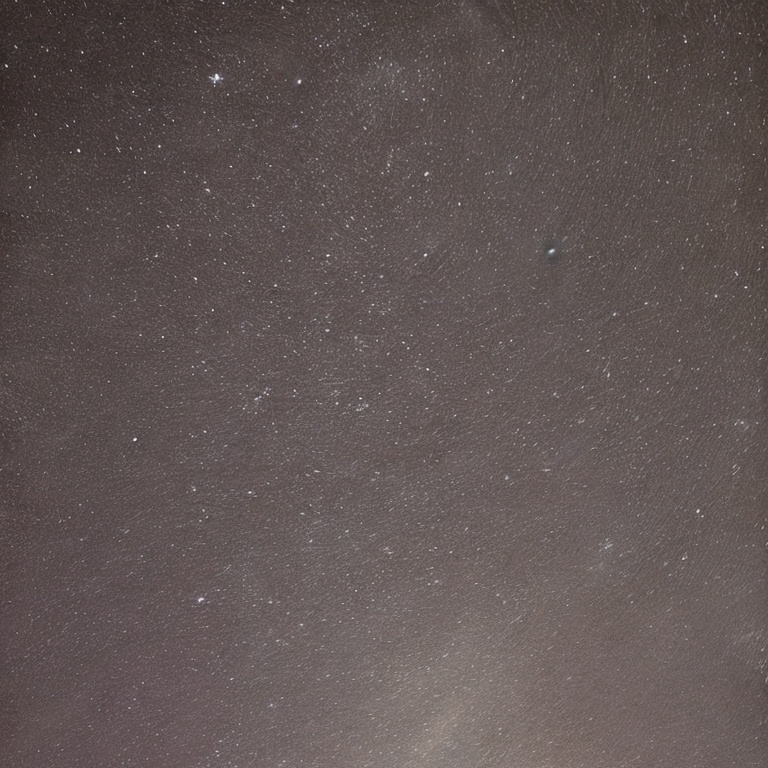}};
            \node [right=0 of img1] (img2) {\includegraphics[width=0.23\linewidth]{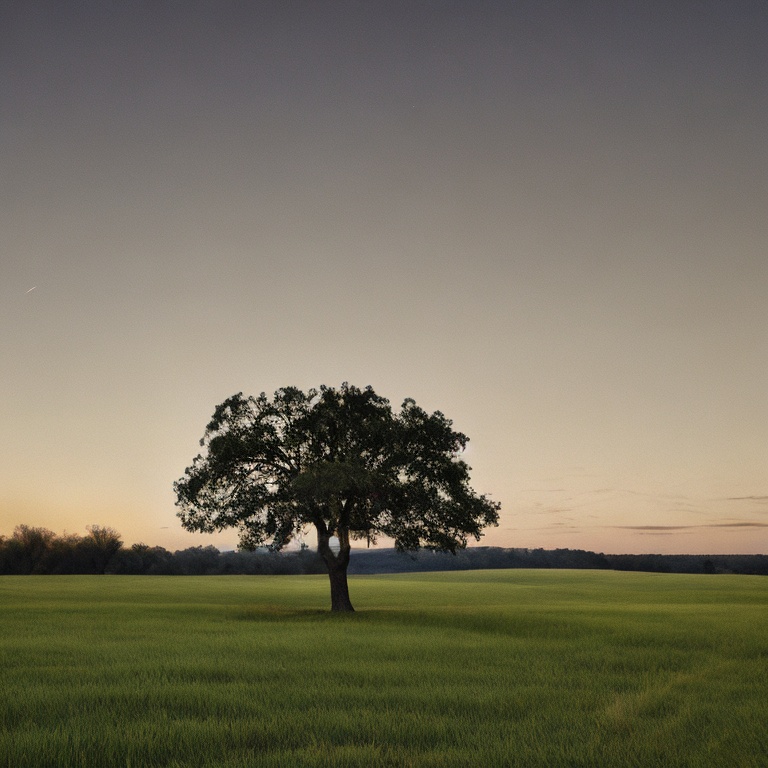}};
            \node [below=0 of img1] (img3) {\includegraphics[width=0.23\linewidth]{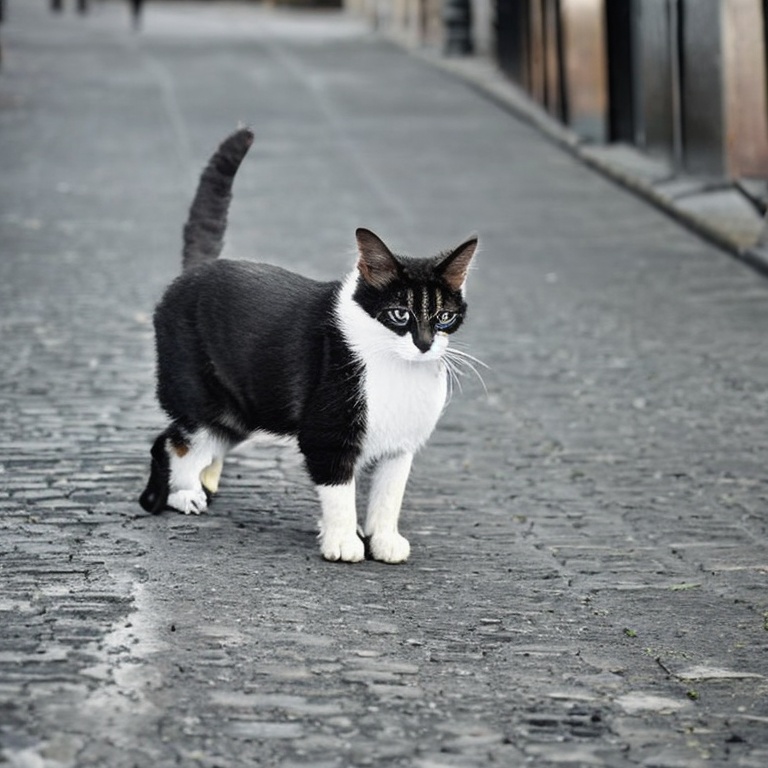}};
            \node [right=0 of img3] (img4) {\includegraphics[width=0.23\linewidth]{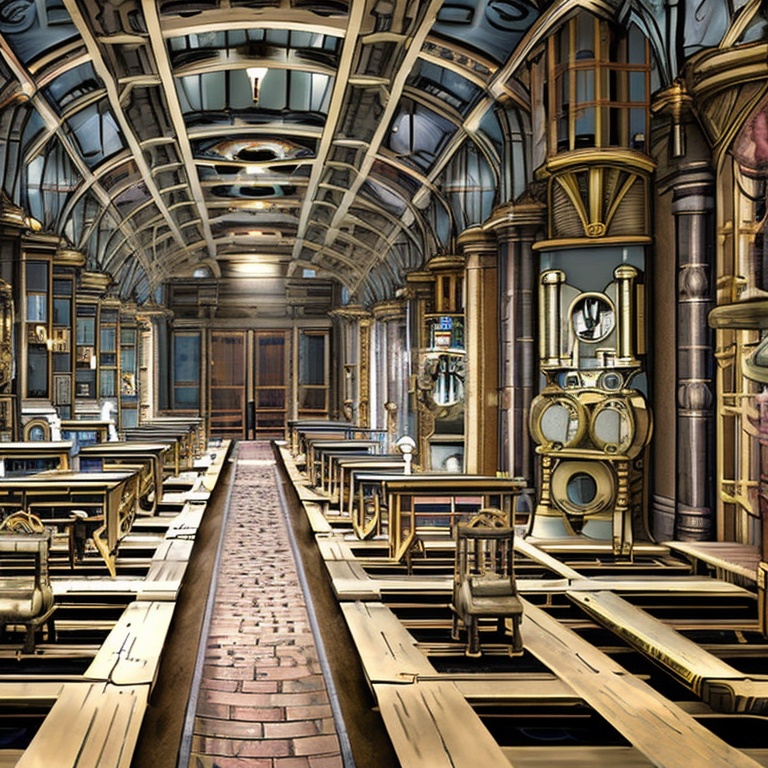}};
            
            \node [right=0.2 of img2] (img11) {\includegraphics[width=0.23\linewidth]{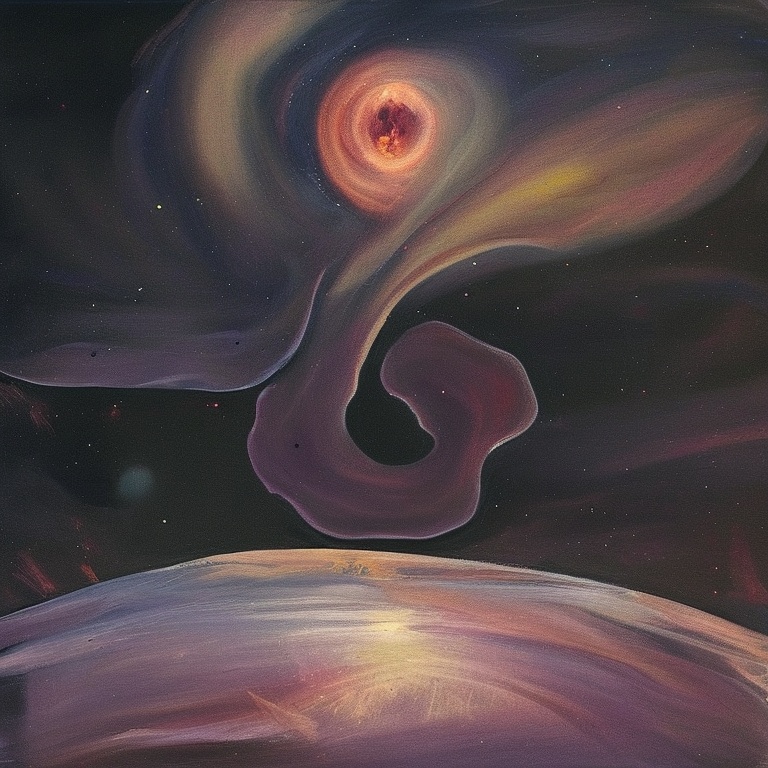}};
            \node [right=0 of img11] (img12) {\includegraphics[width=0.23\linewidth]{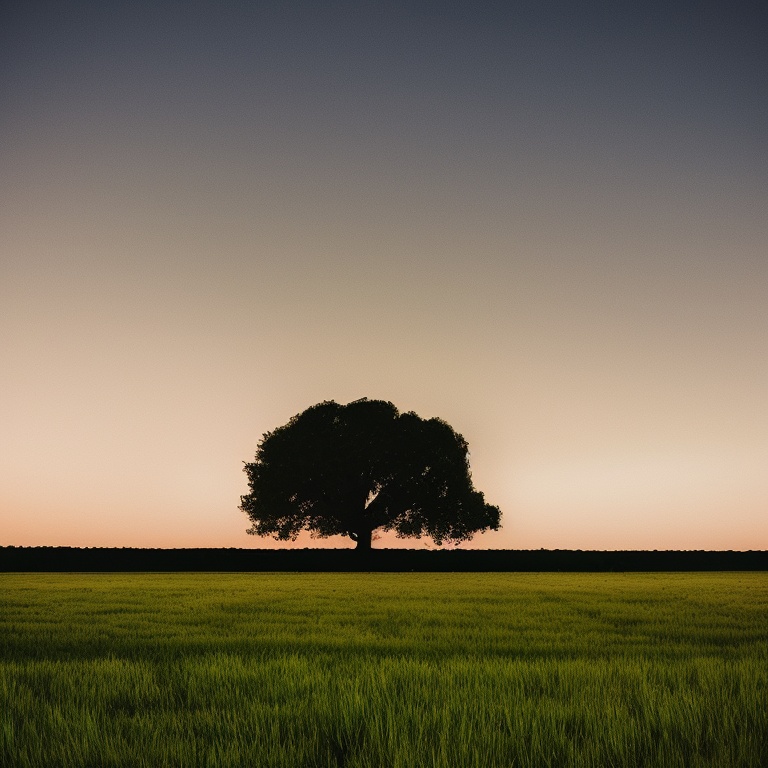}};
            \node [below=0 of img11] (img13) {\includegraphics[width=0.23\linewidth]{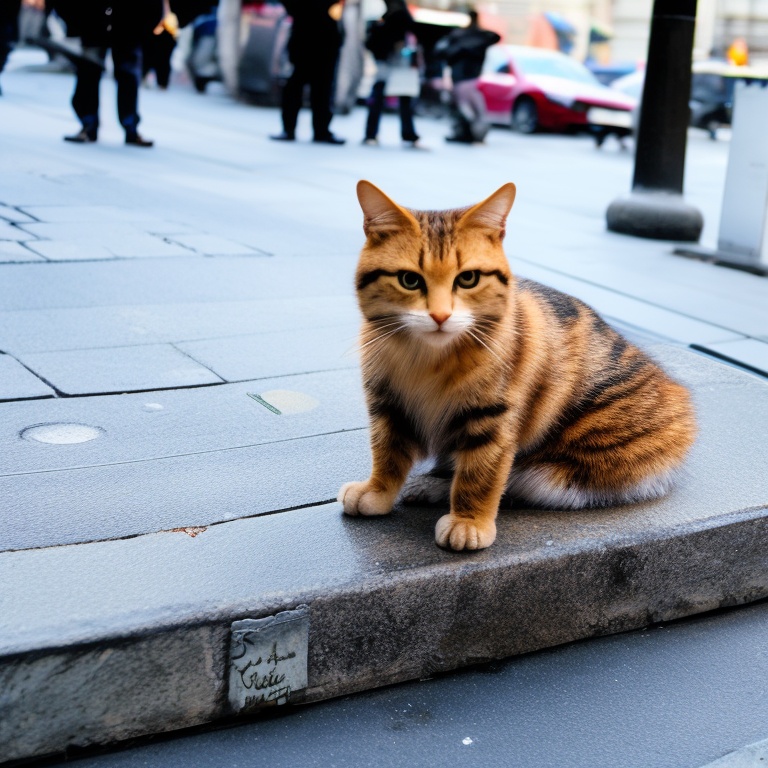}};
            \node [right=0 of img13] (img14) {\includegraphics[width=0.23\linewidth]{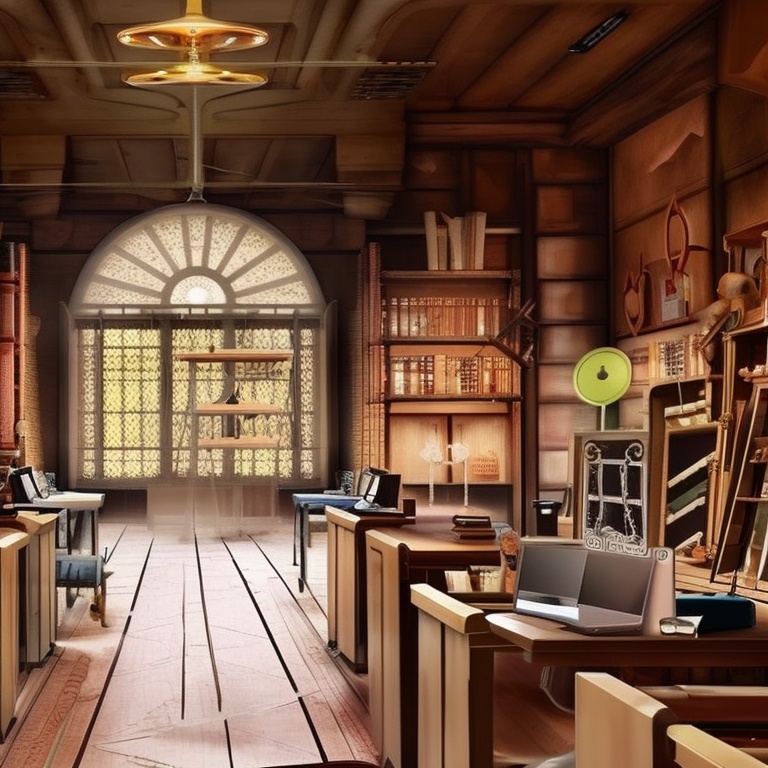}};
            
            \node [fit=(img1) (img2)] (ddim_box) {};
            \node [above=0.15 of ddim_box] {DDIM};
            
            \node [fit=(img11) (img12)] (ddpm_box) {};
            \node [above=0.15 of ddpm_box] {DDPM};
        \end{tikzpicture}
        \caption{Samples of \ac{sd2} generated with \ac{agd} using the \acs{ddpm} sampler, with adapters trained on DDIM trajectories. \Ac{agd} successfully generates high-quality samples when applied with different schedulers at inference.}
        \label{fig:scheduler}
    \end{figure}
}

\newcommand{\IPAdapterSamples}{
    \begin{figure}[t]
        \centering
        \begin{tikzpicture}[every node/.style={inner sep=0pt, outer sep=0pt}]
            \node (img1) {\includegraphics[width=0.3\linewidth]{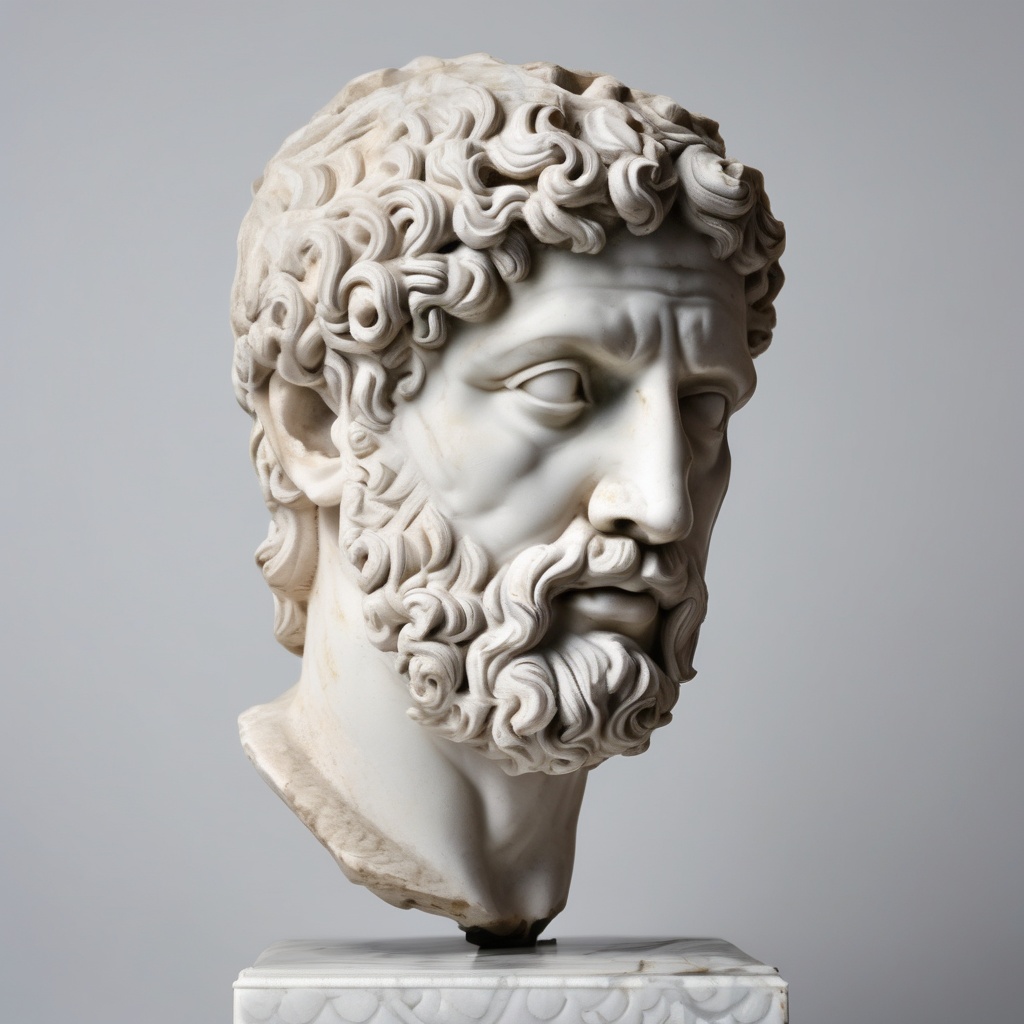}};
            \node [right=0.2 of img1] (img2) {\includegraphics[width=0.3\linewidth]{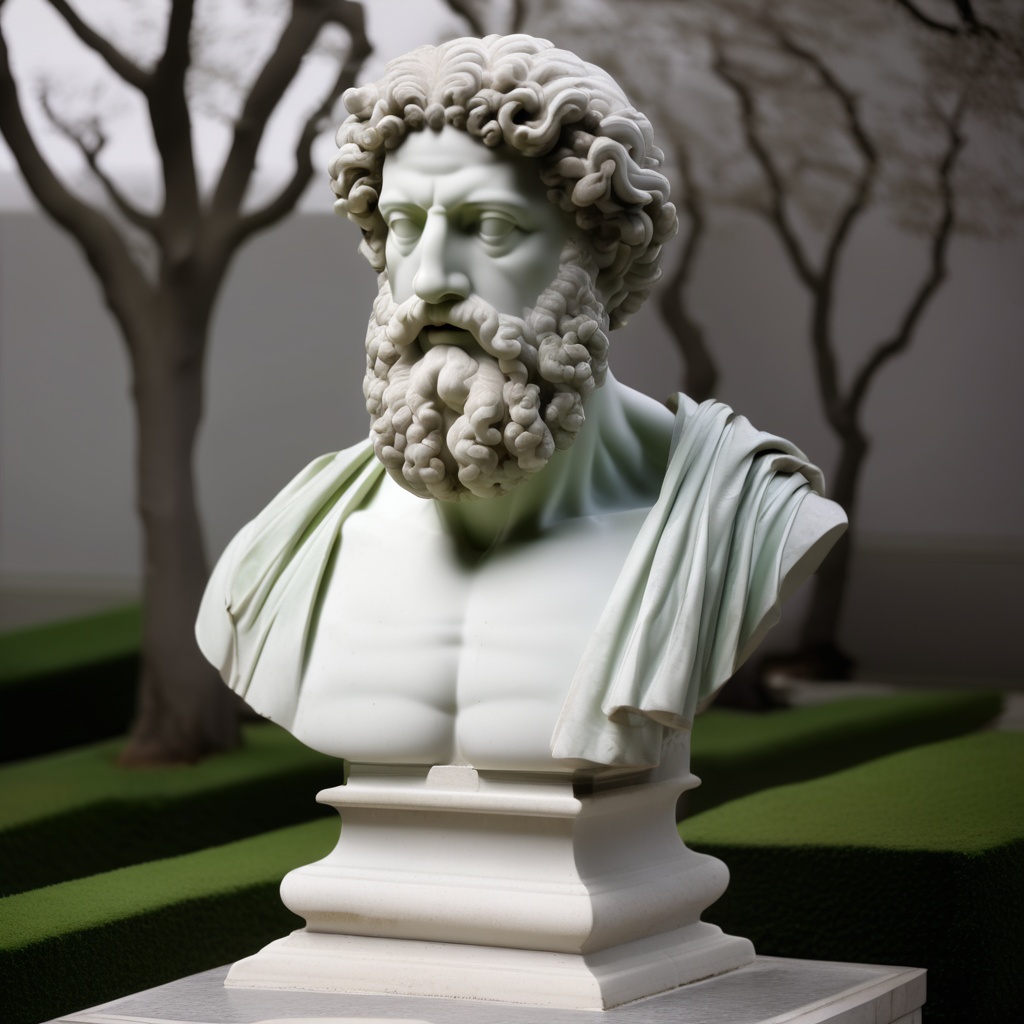}};
            \node [right=0.2 of img2] (img3) {\includegraphics[width=0.3\linewidth]{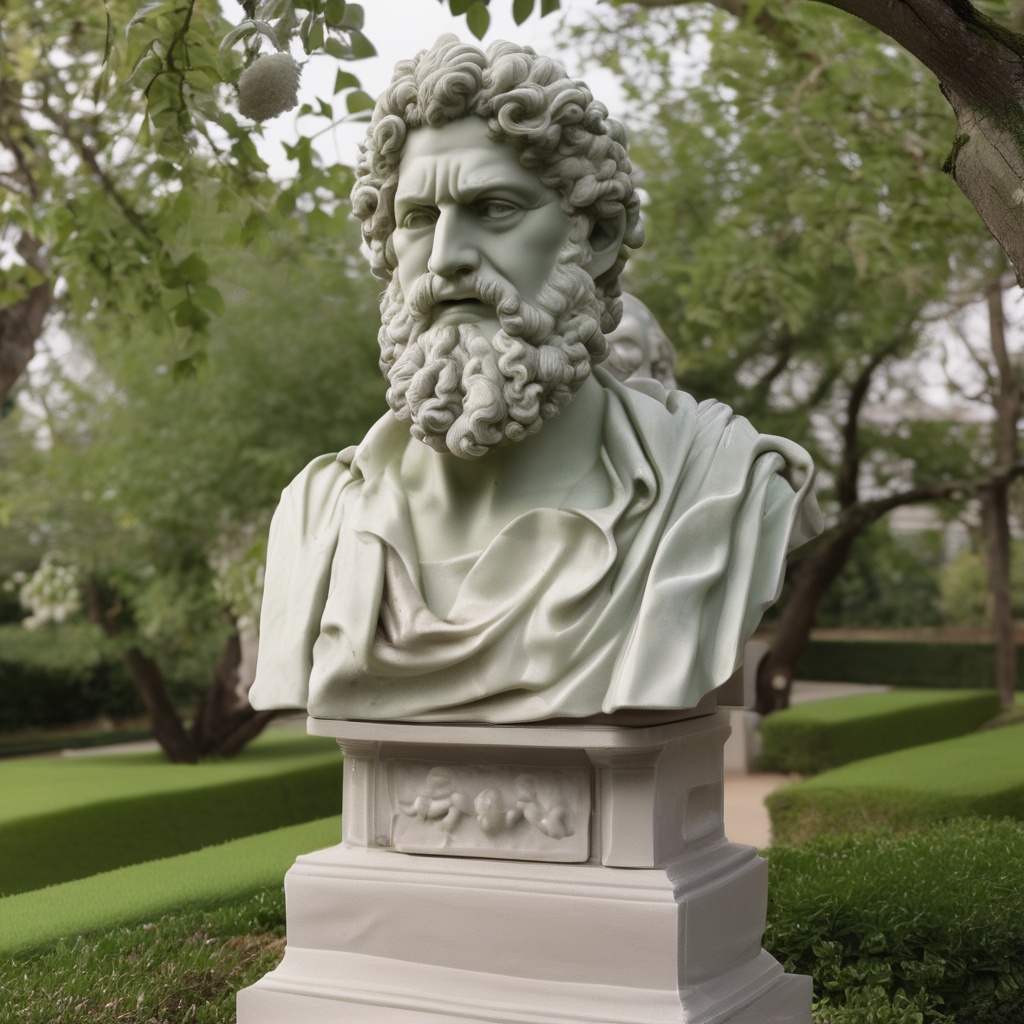}};

            \node [below=0.2 of img1] (img4) {\includegraphics[width=0.3\linewidth]{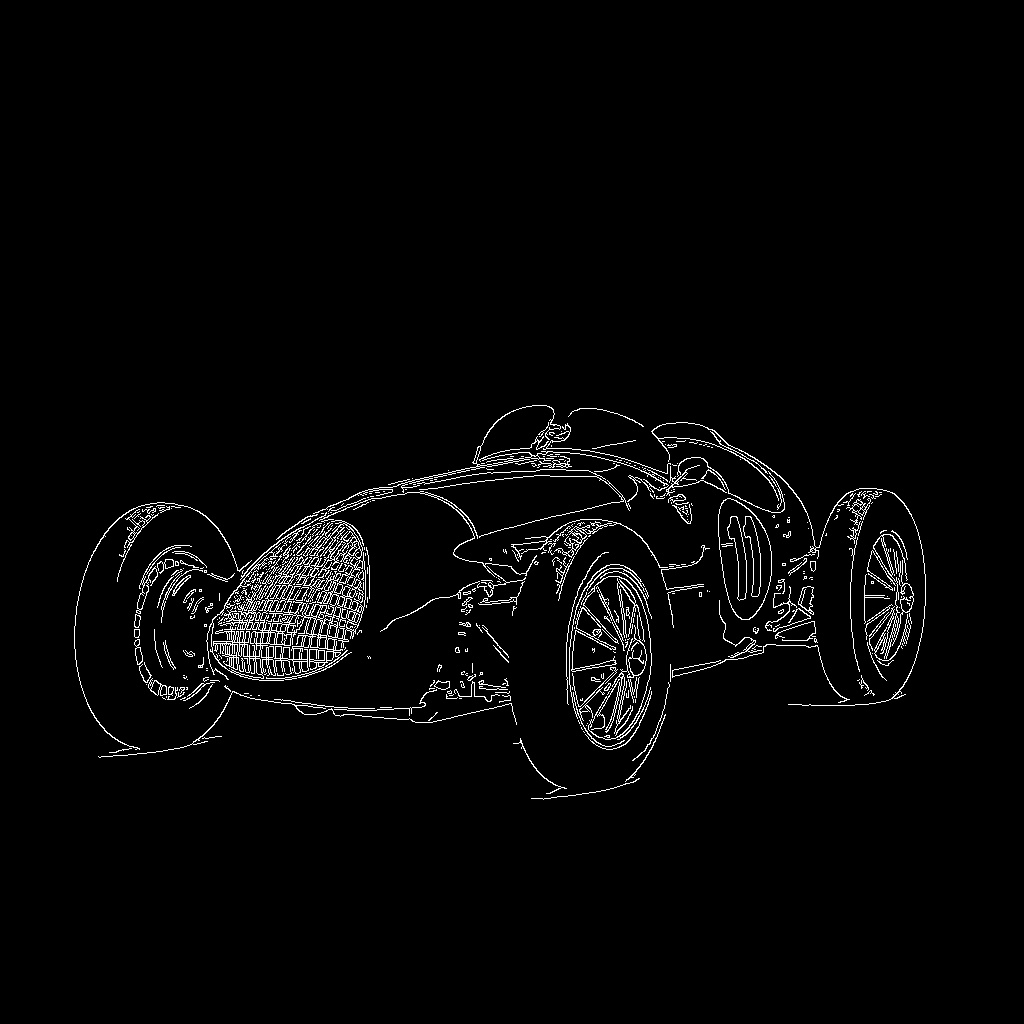}};
            \node [right=0.2 of img4] (img5) {\includegraphics[width=0.3\linewidth]{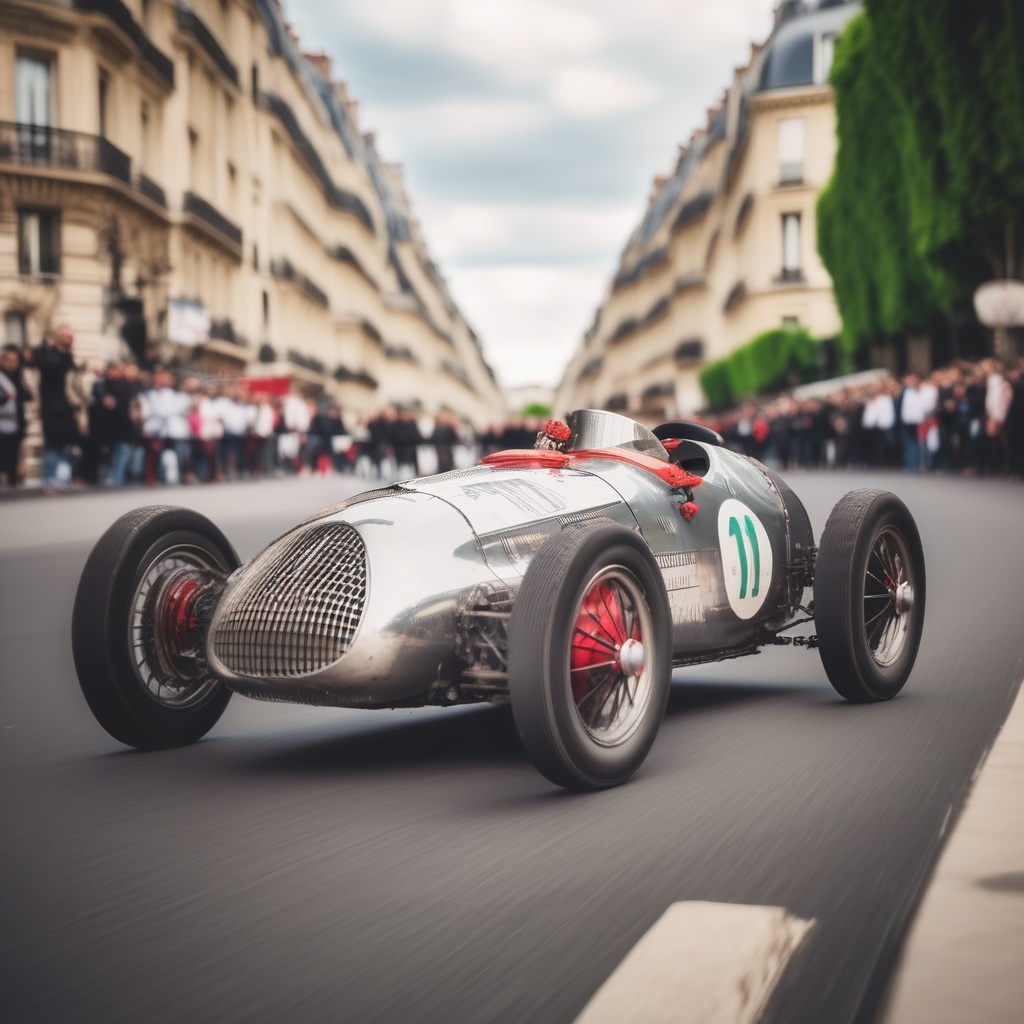}};
            \node [right=0.2 of img5] (img6) {\includegraphics[width=0.3\linewidth]{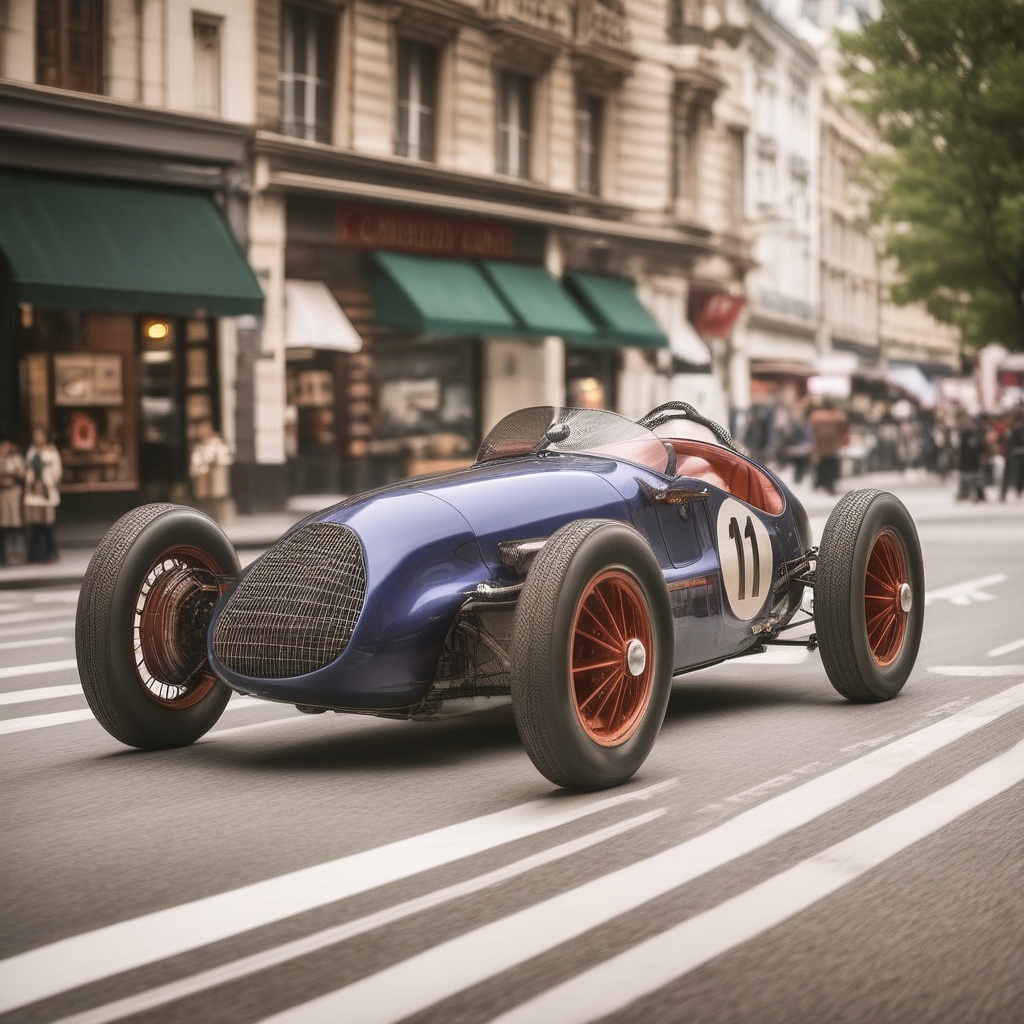}};

            \node [above=0.15 of img1] {Image prompt};
            \node [above=0.15 of img2] {\Acs{cfg}};
            \node [above=0.15 of img3] {\Acs{agd} (Ours)};
        \end{tikzpicture}
        \caption{Using \Ac{agd} with IP-adapter \citep{ye2023ip} and ControlNet \citep{zhang2023adding} for \ac{sdxl}. \ac{agd} can be integrated with other checkpoints derived from the same base model, achieving the benefits of both modules.}
        \label{fig:ip_adapter}
    \vspace{-0.2cm}
    \end{figure}
}

\newcommand{\TrainAlgorithm}{
    \begin{algorithm}[t]
        \caption{Trajectory collection for \ac{agd}.}
        \label{alg:trajectories}
        \begin{algorithmic}[1]
            \Require Set of conditions $\mathcal{C}$
            \Require Guidance scale range $[\omega_{\min}, \omega_{\max}]$
            \Require Pre-trained diffusion model $\vec{\epsilon}_{\vec{\theta}}$
    
            \State $\Omega = \emptyset$
            \For{$c \in \mathcal{C}$}
                \State $\omega \sim \mathrm{Unif}([\omega_{\min}, \omega_{\max}])$
                \State Run the reverse diffusion process from \Cref{eq:diffusion-ode} using $\vec{\epsilon}_{\vec{\theta}}$ with $\omega$ and $c$ in $N$ steps 
                \State Cache $(\vec{x}_t, t, c, \omega)$ and the \ac{cfg}-guided prediction $\tilde{\vec{\epsilon}}_{\vec{\theta}}(\vec{x}_t, t, c, \omega)$ at each sampling step:

                $\Omega \gets \Omega \cup \{ (\vec{x}_t, t, c, \omega, \tilde{\vec{\epsilon}}_{\vec{\theta}}(\vec{x}_t, t, c, \omega)) \}_{t=1}^N$
            \EndFor
    
        \end{algorithmic}
    \end{algorithm}
    \begin{algorithm}[t]
        \caption{Adapter training for \ac{agd}.}
        \label{alg:training}
        \begin{algorithmic}[1]
            \Require Trajectory dataset $\Omega$ from \Cref{alg:trajectories}
            \Require Model with adapters $\vec{\epsilon}_{[\vec{\theta}, \vec{\psi}]}$ 
            \Require Loss function $\ell$
    
            \While{not converged}
                \State $(\vec{x}_t, t, c, \omega, \tilde{\vec{\epsilon}}) \sim \mathrm{Unif}(\Omega)$
                \State $\mathcal{L} = \ell(\tilde{\vec{\epsilon}}, \vec{\epsilon}_{[\vec{\theta}, \vec{\psi}]}(\vec{x}_t, t, c, \omega))$
                \State Update $\vec{\psi}$ with gradient step on $\mathcal{L}$
            \EndWhile
        \end{algorithmic}
    \end{algorithm}
    
}

\newcommand{\Metrics}{%
\begin{table}[t]
    \centering
    \caption{Quantitative comparison between \ac{agd} and \ac{cfg}. \ac{agd} outperforms \ac{cfg} in class-conditional generation using \ac{dit} and performs similarly for text-to-image models (\ac{sd2} and \ac{sdxl}).}
    \label{tab:quantitative}
    \begin{tabular}{
        l
        l
        S[table-format=1.2]
        S[table-format=1.2]
        S[table-format=1.2]
    }
        \toprule
        {\textbf{Model}}
        & {\textbf{Guidance}}
        & {\textbf{FID} $\downarrow$}
        & {\textbf{Precision} $\uparrow$}
        & {\textbf{Recall} $\uparrow$} \\
        \midrule
        \multirow[c]{3}{*}{\Acs{dit}}
        & Unguided & 12.57 & 0.67 & \bfseries 0.74 \\
        & \Acs{cfg} \citep{ho2022classifier} & 5.30 & \bfseries 0.83 & 0.66 \\
        & \mycc \Acs{agd} (Ours) & \mycc \bfseries 5.03 & \mycc 0.80 & \mycc 0.68 \\
        \midrule
        \multirow[c]{3}{*}{\Acs{sd2}}
        & Unguided & 49.94 & 0.39 & \bfseries 0.63 \\
        & \Acs{cfg} \citep{ho2022classifier} & \bfseries 20.94 & \bfseries 0.67 & 0.55 \\
        & \mycc \Acs{agd} (Ours) & \mycc 21.09 & \mycc 0.66 & \mycc 0.55 \\
        \midrule
        \multirow[c]{3}{*}{\Acs{sdxl}}
        & Unguided & 60.30 & 0.35 & \bfseries 0.54 \\
        & \Acs{cfg} \citep{ho2022classifier} & \bfseries 22.82 & 0.66 & 0.52 \\
        & \mycc \Acs{agd} (Ours) & \mycc 22.98 & \mycc \bfseries 0.67 & \mycc 0.52 \\
        \bottomrule
    \end{tabular}
\end{table}%
}

\newcommand{\DiTDropout}{%
\begin{table}[t]
    \caption{Ablation on the dropout rate using  \acs{dit}.}
    \label{tab:dropout}
    \centering
    \begin{tabular}{
        c
        S[table-format=1.2]
        S[table-format=1.2]
        S[table-format=1.2]
    }
        \toprule
        {\textbf{Dropout}}
        & {\textbf{FID $\downarrow$}}
        & {\textbf{Precision $\uparrow$}}
        & {\textbf{Recall $\uparrow$}} \\
        \midrule
        0\% & \bfseries 5.22 & 0.80 & \bfseries 0.67 \\
        10\% & 5.27 & 0.82 & \bfseries 0.67 \\
        20\% & 5.39 & 0.83 & 0.66 \\
        50\% & 5.69 & \bfseries 0.85 & 0.63 \\
        \bottomrule
    \end{tabular}
\end{table}%
}

\newcommand{\DiTHidden}{%
\begin{table}[t]
    \centering
    \caption{Ablation on the hidden dimensionality of the adapters.}
    \label{tab:hidden}
    \begin{tabular}{
        S[table-format=3]
        S[table-format=2.1\%]
        S[table-format=1.2]
        S[table-format=1.2]
        S[table-format=1.2]
    }
        \toprule
        {\textbf{Dim\onedot}}
        & {\textbf{Params}}
        & {\textbf{FID $\downarrow$}}
        & {\textbf{Precision $\uparrow$}}
        & {\textbf{Recall $\uparrow$}} \\
        \midrule
        64 & 0.8\% & 5.33 & \bfseries 0.81 & 0.67 \\
        128 & 2.5\% & \bfseries 5.03 & 0.80 & \bfseries 0.68 \\
        256 & 6.1\% & 5.22 & 0.80 & 0.67 \\
        512 & 17.2\% & 5.26 & \bfseries 0.81 & 0.67 \\
        \bottomrule
    \end{tabular}
\end{table}%
}

\newcommand{\DiTLoss}{%
\begin{table}[t]
    \centering
    \caption{Effect of using different loss functions for distillation.}
    \label{tab:loss}
    \begin{tabular}{
        c
        c
        S[table-format=1.2]
    }
        \toprule
        {\textbf{Loss}}
        & {\textbf{Weight $\lambda(t)$}}
        & {\textbf{FID $\downarrow$}} \\
        \midrule
        $\ell_1$ & 1 & 9.91 \\
        $\ell_2$ & 1 & \bfseries 5.03 \\
        $\ell_2$ & $\sigma(t)$ & 5.30 \\
        $\ell_2$ & $\frac{1}{2} \abs{1 - \cos\angle(\tilde{\vec{\epsilon}}_{\vec{\theta}}, \vec{\epsilon}_{\vec{\theta}})}$ & 6.64 \\
        \bottomrule
    \end{tabular}
\end{table}%
}

\newcommand{\DiTArch}{%
\begin{table}[t]
    \caption{Ablation on adapter architecture using \acs{dit}.}
    \label{tab:arch}
    \centering
    \begin{tabular}{
        l
        S[table-format=1.2]
        S[table-format=1.2]
        S[table-format=1.2]
    }
        \toprule
        {\textbf{Architecture}}
        & {\textbf{FID $\downarrow$}}
        & {\textbf{Precision $\uparrow$}}
        & {\textbf{Recall $\uparrow$}} \\
        \midrule
        Cross-attention & 5.49 & \bfseries 0.83 & 0.65 \\
        Offset & \bfseries 5.03 & 0.80 & \bfseries 0.68 \\
        Positional encoding & 5.25 & 0.81 & 0.66 \\
        Gating & 5.54 & \bfseries 0.83 & 0.66 \\
        \bottomrule
    \end{tabular}
\end{table}%
}

\newcommand{\SDArch}{%
\begin{table}[t]
    \caption{Ablation on the adapter architecture for \acs{sd2}.}
    \label{tab:sd2_arch}
    \centering
    \begin{tabular}{
        l
        S[table-format=1.2]
        S[table-format=1.2]
        S[table-format=1.2]
    }
        \toprule
        {\textbf{Architecture}}
        & {\textbf{FID $\downarrow$}}
        & {\textbf{Precision $\uparrow$}}
        & {\textbf{Recall $\uparrow$}} \\
        \midrule
        Cross-attention & \bfseries 21.09 & \bfseries 0.66 & \bfseries 0.55 \\
        Offset & 22.05 & 0.63 & 0.54 \\
        \bottomrule
    \end{tabular}
\end{table}%
}

\newcommand{\DiTInit}{%
\begin{table}[t]
    \centering
    \caption{Ablation on the initialization type of the adapter layers.}
    \label{tab:init}
    \begin{tabular}{
        l
        S[table-format=1.2]
        S[table-format=1.2]
        S[table-format=1.2]
    }
        \toprule
        {\textbf{Init\onedot scheme}}
        & {\textbf{FID $\downarrow$}}
        & {\textbf{Precision $\uparrow$}}
        & {\textbf{Recall $\uparrow$}} \\
        \midrule
        Zero & 5.24 & \bfseries 0.81 & \bfseries 0.68 \\
        Xavier & \bfseries 5.03 & 0.80 & \bfseries 0.68 \\
        \bottomrule
    \end{tabular}
\end{table}%
}

\newcommand{\DiTAGDDiffusionGD}{%
\begin{table}[t]
    \centering
    \caption{Comparing \ac{agd} and \ac{gd} \citep{meng2023distillation} using \ac{dit} under the same training setup. \Ac{agd} slightly outperforms \ac{gd} while only training the adapters instead of tuning the full model.}
    \label{tab:dit_agd_diffusion_gd}
    \begin{tabular}{
        l
        c
        S[table-format=1.1]
        S[table-format=1.1]
        S[table-format=1.1]
    }
        \toprule
        {\textbf{Method}}
        & {\textbf{Params}}
        & {\textbf{FID} $\downarrow$}
        & {\textbf{Precision} $\uparrow$}
        & {\textbf{Recall} $\uparrow$} \\
        \midrule
        \acs{gd} \citep{meng2023distillation} & 676\,M & 5.66 & \bfseries 0.80 & 0.67 \\
        \mycc \acs{agd} (ours) & \mycc 16\,M & \mycc \bfseries 5.03 & \mycc \bfseries 0.80 & \mycc \bfseries 0.68 \\
        \bottomrule
    \end{tabular}    
    \vspace{-0.2cm}
\end{table}%
\begin{table}[t]
    \centering
    \caption{Importance of training on guided trajectories. \Ac{agd} performs best when trained on \ac{cfg}-guided trajectories instead of the standard diffusion trajectories used in \citep{meng2023distillation}.}
    \label{tab:trajectories}
    \begin{tabular}{
        l
        S[table-format=2.2]
        S[table-format=1.2]
        S[table-format=1.2]
    }
        \toprule
        {\textbf{Guidance method}}
        & {\textbf{FID} $\downarrow$}
        & {\textbf{Precision} $\uparrow$}
        & {\textbf{Recall} $\uparrow$} \\
        \midrule
        \ac{cfg} \citep{ho2022classifier} & 5.30 & \bfseries 0.83 & 0.66 \\
        \mycc \acs{agd} (Diffusion) & \mycc 5.54 & \mycc 0.80 & \mycc \bfseries 0.68 \\
        \mycc \acs{agd} (Trajectory) & \mycc \bfseries 5.03 & \mycc 0.80 & \mycc \bfseries 0.68 \\
        \bottomrule
    \end{tabular}
    \vspace{-0.3cm}
\end{table}
}

\newcommand{\SDScheduler}{%
\begin{table}[t]
    \centering
    \vspace{-0.1cm}
    \caption{Effect of using a different diffusion sampler at inference. The adapter in this case was trained on DDIM trajectories, but other sampling methods such as DDPM can be used at inference.}
    \label{tab:scheduler}
    \begin{tabular}{
        l
        S[table-format=1.2]
        S[table-format=1.2]
        S[table-format=1.2]
    }
        \toprule
        {\textbf{Sampler}}
        & {\textbf{FID} $\downarrow$}
        & {\textbf{Precision} $\uparrow$}
        & {\textbf{Recall} $\uparrow$}
        \\
        \midrule
        DDIM \citep{song2020denoising} & \bfseries 21.09 & 0.66 & \bfseries 0.55 \\
        DDPM \citep{ho2020denoising} & 22.15 & \bfseries 0.67 & 0.51 \\
        \bottomrule
    \end{tabular}
    \vspace{-0.3cm}
\end{table}%
}

\newcommand{\MemoryTrainingSpeed}{%
\begin{table}[t]
    \centering
    \vspace{-0.2cm}
    \caption{Comparing the memory requirements and training speed of \ac{agd} and \ac{gd}. \ac{agd} enables the distillation of large models on an RTX 4090 with 24\,GB of VRAM while also being significantly faster at each training iteration.}
    \label{tab:memory_training_speed}
    \begin{tabular}{
        l
        l
        S[table-format=1.2]
        S[table-format=1.2]
    }
        \toprule
        {\textbf{Model}}
        & {\textbf{Method}}
        & {\textbf{VRAM (GB)}}
        & {\textbf{Speed (it/s)}} 
        \\
        \midrule
        \multirow[c]{2}{*}{\Acs{dit}}
        & \Acs{gd} \citep{meng2023distillation} & 17.67 & 0.52 \\
        & \mycc \Acs{agd} (Ours) & \mycc 16.79 & \mycc 2.36 \\
        \midrule
        \multirow[c]{2}{*}{\Acs{sd2}}
        & \Acs{gd} \citep{meng2023distillation} & \multicolumn{2}{c}{\textit{Out-of-memory}} \\
        & \mycc \Acs{agd} (Ours) & \mycc  23.83 & \mycc 2.05 \\
        \midrule
        \multirow[c]{2}{*}{\Acs{sdxl}}
        & \Acs{gd} \citep{meng2023distillation} & \multicolumn{2}{c}{\textit{Out-of-memory}} \\
        & \mycc \Acs{agd} (Ours) & \mycc 22.77 & \mycc  3.94 \\
        \bottomrule
    \end{tabular}
    \vspace{-0.3cm}
\end{table}%
}

\definecolor{iccvblue}{rgb}{0.21,0.49,0.74}
\usepackage[pagebackref,breaklinks,colorlinks,allcolors=iccvblue]{hyperref}

\def\paperID{10181}
\def\confName{ICCV}
\def\confYear{2025}

\title{Efficient Distillation of Classifier-Free Guidance using Adapters}

\author{Cristian Perez Jensen$^{*}$  \quad  Seyedmorteza Sadat$^{*}$ \vspace{0.2cm}\\
ETH Z\"urich\\
{\tt\small \{cjense, ssadat\}@ethz.ch}
}

\begin{document}

\twocolumn[{%
    \maketitle%
    \vspace{-.7cm}
    \Teaser%
    \vspace{.4cm}
}]

\maketitle

\acresetall
\def\thefootnote{*}\footnotetext{These authors contributed equally to this work}

\begin{abstract}
While \ac{cfg} is essential for conditional diffusion models, it doubles the number of \acp{nfe} per inference step. To mitigate this inefficiency, we introduce \ac{agd}, a novel approach that simulates \ac{cfg} in a single forward pass. \ac{agd} leverages lightweight adapters to approximate \ac{cfg}, effectively doubling the sampling speed while maintaining or even improving sample quality. Unlike prior guidance distillation methods that tune the entire model, \ac{agd} keeps the base model frozen and only trains minimal additional parameters ($\sim$2\%) to significantly reduce the resource requirement of the distillation phase. Additionally, this approach preserves the original model weights and enables the adapters to be seamlessly combined with other checkpoints derived from the same base model. We also address a key mismatch between training and inference in existing guidance distillation methods by training on \ac{cfg}-guided trajectories instead of standard diffusion trajectories. Through extensive experiments, we show that \ac{agd} achieves comparable or superior \acs{fid}  to \ac{cfg} across multiple architectures with only half the \acp{nfe}. Notably, our method enables the distillation of large models ($\sim$2.6B parameters) on a single consumer GPU with 24\,GB of VRAM, making it more accessible than previous approaches that require multiple high-end GPUs. We will publicly release the implementation of our method.
\end{abstract}    
\acresetall
\section{Introduction} \label{sec:intro}

Score-based diffusion models \citep{sohl2015deep,ho2020denoising,song2020score} are a family of generative models that learn the data distribution by reversing a forward process that progressively corrupts the data until it becomes indistinguishable from pure Gaussian noise. Theoretically, running the reverse diffusion process should enable accurate sampling from the data distribution, assuming access to the ground truth score function. However, in practice, unguided sampling from diffusion models often produces low-quality images that fail to align well with the given input condition due to optimization errors \citep{karras2024guiding}. Accordingly, \ac{cfg} \citep{ho2022classifier} has become a crucial technique in modern conditional diffusion models to enhance both generation quality and alignment to conditioning signals---though this comes at the expense of reduced sample diversity \citep{ho2022classifier,sadat2024cads}.

\Ac{cfg} is an inference method that enhances generation quality by leveraging the difference between conditional and unconditional model predictions at each inference step. This difference serves as an update direction to improve both quality and alignment with the target condition. However, \ac{cfg} requires two forward passes per inference step, resulting in twice the sampling cost compared to unguided sampling. This increased cost introduces a significant computational overhead, especially when sampling from large-scale diffusion models or employing these pre-trained models for tasks such as score distillation \citep{poole2023dreamfusion}.

In this paper, we aim to double the sampling speed of \ac{cfg} by training a small set of adapters to integrate the \ac{cfg} behavior directly into the model. Our method, called \ac{agd}, learns to replicate the \ac{cfg}-guided output at each inference step using a single forward pass while preserving the original diffusion model weights. These lightweight adapters add only 1--5\% more parameters to the base model and introduce negligible latency overhead during inference. Since the base model remains frozen during training, and only the adapter parameters are updated, \ac{agd} is computationally efficient and can be trained on a single consumer GPU with 24\,GB of VRAM, even for large models like \ac{sdxl}. Furthermore, \ac{agd} allows the trained adapters to be seamlessly integrated with other checkpoints originating from the same base model, such as IP-adapters \citep{ye2023ip}. We demonstrate that our approach maintains or improves generation quality compared to standard \ac{cfg} and outperforms existing methods such as \acf{gd} \citep{meng2023distillation}, all while significantly reducing resource requirements during training.   

Moreover, we identify and address a mismatch between training and inference trajectories in prior guidance distillation methods. We argue that effective guidance distillation requires training on \emph{\ac{cfg}-guided trajectories} computed by running the sampling process with CFG, as these differ significantly from standard diffusion trajectories obtained by adding noise to the training data. Furthermore, training on guided trajectories eliminates the need to load a teacher model during distillation, thus reducing memory requirements when training \ac{agd}. Another advantage is that the distillation can be performed entirely on the synthetic data generated by the teacher model without needing any real dataset in advance. Our experiments demonstrate that training on \ac{cfg}-guided trajectories enhances performance compared with training on diffusion trajectories.

\Cref{fig:trajectory_training_inference} gives an overview of different components in \ac{agd}. In summary, our primary contributions are as follows:
\begin{itemize}
    \item We introduce \ac{agd}, an efficient method for simulating \ac{cfg} in a single forward pass by training lightweight adapters alongside a frozen base diffusion model, eliminating the need to fine-tune the entire model.
    
    \item We propose training \ac{agd} on \ac{cfg}-guided trajectories instead of diffusion trajectories, reducing the mismatch between training and inference and improving performance.
    
    \item We demonstrate the resource efficiency of \ac{agd} by successfully distilling \ac{sdxl} (2.6B parameters) on a single RTX 4090 GPU with 24\,GB of VRAM.
    
    \item Through extensive experiments, we show that \ac{agd} matches or surpasses \ac{cfg} in performance across various models such as \acl{dit} and Stable Diffusion while doubling the sampling speed compared to \ac{cfg}.
\end{itemize}

\TrainingInference
\section{Related work}

Diffusion models \citep{sohl2015deep, song2019generative, song2020denoising, song2020score, ho2020denoising} have emerged as a leading approach for generative modeling across various domains, including images \citep{ramesh2022hierarchical, rombach2022high, dai2023emu, saharia2022photorealistic}, text \citep{hoogeboom2021argmax, li2022diffusion, austin2021structured}, audio \citep{evans2024fast}, and molecular generation \citep{hoogeboom2022equivariant}. Since the introduction of \acs{ddpm} \citep{ho2020denoising}, significant progress has been made in multiple aspects, such as refining network architectures \citep{hoogeboom2023simple,karras2024analyzing,peebles2023scalable,dhariwal2021diffusion}, developing more efficient sampling methods \citep{song2020denoising,karras2022elucidating,plms,dpm_solver,salimansProgressiveDistillationFast2022}, and introducing novel training strategies \citep{nichol2021improved,song2020score,salimans2022progressive,rombach2022high,karras2022elucidating,karras2024analyzing}. Nevertheless, various guidance techniques \citep{dhariwal2021diffusion,ho2022classifier,karras2024guiding,sadat2025no} have remained essential for enhancing generation quality and the alignment between conditioning inputs and generated outputs \citep{glide}, though they lead to increased sampling time \citep{ho2022classifier} and reduced diversity \citep{sadat2024cads,kynkaanniemi2024applying}.

Several works have recently explored modifying the weight schedule of \ac{cfg} by applying guidance only at certain sampling steps \citep{castillo2023adaptive, wang2024analysis,kynkaanniemi2024applying}, primarily to balance diversity and quality in generation. However, these methods still require two \acp{nfe} for most steps and therefore cannot fully double the inference speed. Additionally, our approach is orthogonal to these methods, as the distilled model can be used for the steps where \ac{cfg} is applied in the above works.

Alternatively, \citet{meng2023distillation} introduced \acf{gd}, which fine-tunes a diffusion model to generate guided predictions in a single forward pass. However, fully fine-tuning the base model is often inefficient and unstable for large models, overwrites the original model weights, and demands high-end GPUs with substantial memory for training. To address these limitations, we propose a more efficient distillation method using adapters. Moreover, \citet{meng2023distillation} trained their model on standard diffusion trajectories, which we show to be less effective than training the distillation process on \ac{cfg}-guided trajectories.

Finally, adapters \citep{houlsby2019parameter} have emerged as a parameter-efficient solution for fine-tuning large-scale diffusion models, mainly for integrating image conditions into pre-trained text-to-image models \citep{mou2024t2i, ye2023ip}. In contrast, we leverage adapters to inject guided predictions directly into the model’s forward pass. Notably, we demonstrate that adapters not only require significantly fewer training resources but also slightly outperform full fine-tuning.
\section{Background} \label{sec:background} 

In this section, we provide a concise overview of diffusion models. Consider a data point $\vec{x} \sim \pdata$ and noise $\vec{\epsilon} \sim \mathcal{N}(\vec{0}, \mat{I})$, and let the forward diffusion process be defined as $\vec{x}_t = \vec{x} + \sigma(t) \vec{\epsilon}$, where noise is gradually introduced over time $t \in [0, T]$. The function $\sigma(t)$ serves as the noise schedule, determining the extent of perturbation at each step, with $\sigma(0) = 0$ and $\sigma(T) = \sigma_{\max}$. As shown by \citet{karras2022elucidating}, this process can be described by the following \ac{ode}:  
\begin{equation}\label{eq:diffusion-ode}
    \dd\vec{x}_t = - \dot{\sigma}(t) \sigma(t) \grad_{\vec{x}_t} \log p_t(\vec{x}_t) \dd t,
\end{equation}  
where $p_t$ is the marginal distribution of the noisy data at time step $t$, transitioning from the original data distribution $p_0 = \pdata$ to a Gaussian prior $p_T = \mathcal{N}(\vec{0}, \sigma_{\max}^2 \mat{I})$.  

Assuming access to the time-dependent score function $\nabla_{\vec{x}_t} \log p_t(\vec{x}_t)$, one can solve this \ac{ode} in reverse, \ie, from $t = T$ to $t = 0$, to generate new samples from $\pdata$. The unknown score function $\nabla_{\vec{x}_t} \log p_t(\vec{x}_t)$ is typically learned using a neural denoiser $D_{\vec{\theta}}(\vec{x}_t, t)$, which is trained to recover clean samples $\vec{x}$ from their noisy counterparts $\vec{x}_t$. Additionally, conditional generation can be achieved by extending the denoiser to $D_{\vec{\theta}}(\vec{x}_t, t, c)$, where $c$ represents auxiliary conditioning information, such as class labels or text.  

\paragraph{Training}
Following \citep{ho2020denoising}, the denoiser $\denoiser$ is commonly parameterized as  
\begin{equation}
    D_{\vec{\theta}}(\vec{x}_t, t) = \vec{x}_t - \sigma(t)\vec{\epsilon}_{\vec{\theta}}(\vec{x}_t, t),
\end{equation}  
and is trained by predicting the added noise $\vec{\epsilon}$ in $\vec{x}_t$, that is by solving
\begin{equation}
    \vec{\theta}^\star \in \argmin_{\vec{\theta}} \E_{\vec{x}, \vec{\epsilon}, t} \lft[\|\vec{\epsilon}_{\vec{\theta}}(\vec{x}_t, t) - \vec{\epsilon}\|^2\rgt].
\end{equation}  
After training, the score function can be approximated via  
\begin{equation}
    \grad_{\vec{x}_t} \log p_t(\vec{x}_t) \approx \frac{D_{\vec{\theta}}(\vec{x}_t, t) - \vec{x}_t}{\sigma(t)^2} = -\frac{\vec{\epsilon}_{\vec{\theta}}(\vec{x}_t, t)}{\sigma(t)}.
\end{equation}

\paragraph{\Acl{cfg}}
\Ac{cfg} is an inference technique aimed at improving the quality of generated samples by blending the outputs of a conditional and an unconditional model \citep{ho2022classifier}. Specifically, \ac{cfg} adjusts the denoiser’s output at each sampling step according to
\begin{equation} \label{eq:cfg}
    \predcfg = \wcfg\pred - (\wcfg - 1)\prednull,
\end{equation} 
where $\wcfg = 1$ corresponds to unguided sampling, and $c = \emptyset$ represents the unconditional prediction. The unconditional model $\prednull$ is typically trained by randomly replacing the conditioning input with $c = \emptyset$ during training. Alternatively, a dedicated denoiser can be trained separately to approximate the unconditional score \citep{karras2024analyzing}. \Ac{cfg} is known to significantly improve generation quality, though it comes at the cost of doubling the sampling time \citep{ho2022classifier}.
\section{\Acl{agd}}

We now introduce our method, \acf{agd},  for doubling the sampling speed of \ac{cfg}.  As shown in \Cref{fig:trajectory_training_inference}, \ac{agd} consists of two main components:
\begin{enumerate*}[(1)]
    \item training on \ac{cfg}-guided trajectories instead of standard diffusion trajectories, and
    \item training lightweight adapters to distill \ac{cfg} instead of fine-tuning the full model.
\end{enumerate*}
Below, we discuss each component in detail, with \Cref{alg:trajectories,alg:training} also providing the training details of \ac{agd}.

\subsection{Training on \acs{cfg}-guided trajectories} \label{sec:trajectories}
\TrainAlgorithm

Prior guidance distillation methods are trained on standard diffusion trajectories, where noise is added to the training data, and the \ac{cfg} prediction is matched at each inference step \citep{meng2023distillation}. However, since \ac{cfg} modifies the reverse process of diffusion models, guided trajectories differ significantly from standard diffusion trajectories, as shown in \Cref{fig:trajectory_densities}. We argue that training directly on \ac{cfg}-guided trajectories enhances guidance distillation by exposing the model to regions in space that the guided reverse process will follow. To bridge the gap between training and inference, we thus train \ac{agd} directly on \ac{cfg}-guided trajectories. We generate guided trajectories as outlined in \Cref{alg:trajectories}, which are then used to train \ac{agd}. Since these trajectories can be cached, the teacher model does not need to be loaded during training, freeing up VRAM. Moreover, because this method only makes use of samples of the teacher model, it does not require an external dataset to train. Additionally, the trajectory dataset only needs to be collected once, enabling efficient hyperparameter tuning for the adapters.

\TrajectoryDensities

\subsection{Efficient distillation with adapters} \label{sec:adapters}

For more efficient training, \ac{agd} only uses small learnable modules, or \emph{adapters} \citep{houlsby2019parameter}, to replicate the effect of \ac{cfg}. Unlike tuning the whole diffusion network as in \ac{gd} \citep{meng2023distillation}, we freeze the original model weights, ensuring that the base model is still available after training. This also allows us to use the learned adapters with other checkpoints that are obtained from the same base model, such as IP-adapters \citep{ye2023ip}. The details of the adapters used in \ac{agd} are given below.

\paragraph{Adapter formulation}
Let $f_{\vec{\theta}}$ denote an intermediate layer in the network with $\mat{Z} \in \R^{L \times d}$ being its upstream input. Further, $f_{\vec{\theta}}$ receives the time step $t$ and the condition embedding $c$  as input. An adapter $g_{\vec{\psi}}$ with parameters $\vec{\psi}$ is a layer that combines $f_{\vec{\theta}}$ with encodings of the guidance scale $\omega$, and the input conditions $(t, c)$ via residual connection:
\begin{equation}
    \tilde{f}_{[\vec{\theta},\vec{\psi}]}(\mat{Z}, \omega, t, c) = f_{\vec{\theta}}(\mat{Z}, t, c) + \adapter.
\end{equation}
This architecture is illustrated in \Cref{fig:adapter}. During training, the model weights $\vec{\theta}$ are kept frozen and only the adapter parameters $\vec{\psi}$ are optimized to match the \ac{cfg} step based on the trajectory dataset, as introduced in \Cref{sec:trajectories}, \ie,
\begin{equation}
    \vec{\psi}^\star \in \argmin_{\vec{\psi}} \E\lft[\ell(\vec{\epsilon}_{[\vec{\theta},\vec{\psi}]}(\vec{x}_t, t, c, \omega), \tilde{\vec{\epsilon}}_{\vec{\theta}}(\vec{x}_t, t, c, \omega))\rgt],
\end{equation}
where $\tilde{\vec{\epsilon}}_{\vec{\theta}}(\vec{x}_t, t, c, \omega)$ denotes a \ac{cfg} step with guidance scale $\omega$, $\vec{\epsilon}_{[\vec{\theta},\vec{\psi}]}(\vec{x}_t, t, c, \omega)$ is the output of the model with the adapters, and $\ell$ is the loss function.

\paragraph{Adapter architecture}
We mainly experiment with two adapter architectures:
\begin{enumerate*}[(1)]
    \item cross-attention adapters, and
    \item offset adapters.
\end{enumerate*}
Let $\mat{C} = [\vec{c}_1, \ldots, \vec{c}_C]$ represent the matrix containing all conditioning embeddings (\eg, guidance scale, prompt embeddings, \etc), linearly projected to the same dimensionality via a learned projection. Akin to IP-adapter \citep{ye2023ip}, the cross-attention adapter formulates $g_{\vec{\psi}}$ as
\begin{equation} \label{eq:ca}
    \adapter = \mathrm{Softmax}\prn{\frac{\mat{Q} \trp{\mat{K}}}{\sqrt{d}}} \mat{V},
\end{equation}
where $\mat{Q} = \mat{Z} \mat{W}_q$, $\mat{K} = \mat{C} \mat{W}_k$, and $\mat{V} = \mat{C} \mat{W}_v$. The offset adapter formulates $g_{\vec{\psi}}$ as 
\begin{equation} \label{eq:offset}
    \adapter = \mathrm{MLP}\prn{\sum_{i=1}^C \vec{c}_i}
\end{equation}
We found that offset adapters perform better for simpler models like \ac{dit}, whereas cross-attention adapters are more effective for text-to-image models. Several ablations on the adapter design space are provided in \Cref{app:ablations}.

\AdapterAndBaseModel

\paragraph{Implementation details}
We embed the guidance scale $\omega$ via a Fourier feature encoder \citep{tancik2020fourier} followed by a \ac{mlp}. We also extract the text or class embeddings from the base model (\eg, CLIP embeddings) and linearly project them into the same dimensionality as the guidance scale embedding. In \ac{dit} \citep{peebles2023scalable}, we place an adapter in each transformer block after the self-attention mechanism. For text-to-image models such as \ac{sd2} \citep{rombach2022high} and \ac{sdxl} \citep{podell2024sdxl}, we place the adapters in conjunction with the cross-attention layers, since the text prompt is only used in these blocks.

\paragraph{Efficiency}
Since the adapters introduce only 1--5\% additional parameters relative to the underlying model, their computational overhead remains negligible during both training and inference. Furthermore, unlike \ac{cfg}, which requires \emph{two} forward passes per diffusion step, our approach performs only \emph{one}, effectively halving the \acp{nfe}. Consequently, our method achieves twice the speed of \ac{cfg} when generating samples from pre-trained diffusion models.

\section{Experiments} \label{sec:experiments}

\paragraph{Setup}
We evaluate \ac{agd} on class-conditional generation using \acf{dit} \citep{peebles2023scalable}, and text-to-image generation using \acf{sd2} \citep{rombach2022high} and \acf{sdxl} \citep{podell2024sdxl}. All experiments are conducted on a single RTX 4090 GPU (24\,GB of VRAM). Training is performed using the Adam optimizer \citep{kingma2014adam} without weight decay, where the learning rate follows a linear warm-up to $1 \times 10^{-4}$ over the first 10\% of steps, after which it decays via a cosine annealing schedule \citep{loshchilov2016sgdr}. For training adapters on \ac{dit}, trajectories are sampled with guidance scales $\omega \sim \mathrm{Unif}([1, 6])$, with four trajectories per class label of ImageNet \citep{deng2009imagenet}. For text-to-image models, we randomly select 500 captions from the COCO-2017 training set \citep{lin2014microsoft}, generating a single trajectory per caption with guidance scales $\omega \sim \mathrm{Unif}([1, 12])$. Please refer to \Cref{app:details} for additional details regarding the experiments.

\paragraph{Evaluation metrics}
We mainly use \ac{fid} \citep{heusel2017gans} to measure the quality and diversity of generated images, since it closely aligns with human perception. Given \ac{fid}'s sensitivity to implementation details, we evaluate all models under identical conditions to ensure consistency. Additionally, we report precision as a measure of generation quality and recall as an indicator of diversity \citep{kynkaanniemi2019improved}.

\Samples
\subsection{Qualitative results}
We evaluate the qualitative performance of \ac{agd} and \ac{cfg} in \Cref{fig:samples}, generating samples using the same random seeds for both methods. Our results indicate that \ac{agd} produces images structurally similar to \ac{cfg} while being more visually appealing across multiple models and resolutions. Thus, \ac{agd} retains the quality benefits of \ac{cfg} while achieving twice the sampling speed per image.
\Metrics

\subsection{Quantitative results}
The quantitative evaluation of \ac{agd} and \ac{cfg} is shown in \Cref{tab:quantitative}. We observe that \ac{agd} achieves metrics comparable to \ac{cfg}, with both methods significantly outperforming the unguided sampling baseline. This confirms that \ac{agd} enhances generation quality similarly to \ac{cfg} while requiring only half the \acp{nfe}. Notably, \ac{agd} even slightly outperforms \ac{cfg} for the \ac{dit} model.

\subsection{Comparing \ac{agd} with \acl{gd}} \label{sec:experiments:agd_gd}
\DiTAGDDiffusionGD
We next compare our method to \acf{gd} \citep{meng2023distillation}, which fine-tunes the entire diffusion model to replicate guided outputs. We train \ac{agd} and \ac{gd} under the same training setup using \ac{dit} as the base diffusion model for class-conditional ImageNet generation. \Cref{tab:dit_agd_diffusion_gd} shows that \ac{agd} outperforms \ac{gd} in \ac{fid} while having significantly less trainable parameters. Thus, we conclude that \ac{gd} can be made significantly more efficient by keeping the base model frozen and only training the adapters.

Moreover, \Cref{fig:ood} shows that \ac{gd} completely fails when used with guidance scales outside the domain seen during fine-tuning. In contrast, \ac{agd} remains robust to this issue, demonstrating better generalization across guidance scales.
\SamplesOOD

\subsection{Importance of training on guided trajectories} \label{sec:experiments:guidance_diffusion}
\IPAdapterSamples
To validate our claim that training on guidance trajectories is beneficial, we compare \ac{agd} trained on standard diffusion trajectories with \ac{agd} trained on guidance trajectories. As shown in \Cref{tab:trajectories}, training on guidance trajectories yields a substantial improvement over training on standard diffusion trajectories. Hence, we conclude that bridging the train-inference gap by aligning these trajectories enhances performance, as it focuses training on regions of the space that are important for \ac{cfg}.

\subsection{Training efficiency} \label{app:speed}
\Cref{tab:memory_training_speed} compares the training speed and VRAM usage of \ac{agd} and \ac{gd} \citep{meng2023distillation}. We note that for larger networks like \ac{sdxl}, \ac{agd} can successfully distill the model using a consumer GPU with 24\,GB VRAM, whereas \ac{gd} encounters out-of-memory issues. Even when VRAM is not a constraint, each training step of \ac{agd} remains significantly more efficient than \ac{gd} ($\sim$4.5$\times$ faster for \ac{dit}).
\MemoryTrainingSpeed

\subsection{Combining \ac{agd} with other adapters}
\label{sec:experiments:ip_adapter}
\Cref{fig:ip_adapter} shows samples from combining \ac{agd} with IP-adapter \citep{ye2023ip} and ControlNet \citep{zhang2023adding}. As can be seen, \ac{agd} continues to produce high-quality samples when other adapters or networks are applied to the same model, while achieving the goals of both modules.

\subsection{\Ac{agd} vs guidance scale} \label{sec:experiments:guidance_scale_selection}
\GuidanceVsFID
\Cref{fig:guidance_vs_fid} shows how the performance of \ac{agd} varies as we increase the guidance scale $\wcfg$. We observe that the \ac{fid} curve for \ac{cfg} is more peaked, whereas the curve for \ac{agd} is relatively flatter, making it less sensitive to the exact guidance value at inference for good performance. Additionally, we note that \ac{agd} have a more favorable trade-off between precision and recall compared to \ac{cfg}, resulting in better \ac{fid} scores for most guidance scales.

\subsection{Changing the scheduler} \label{sec:experiments:schedulers}
Next, we show that \ac{agd} is not sensitive to the exact choice of scheduler used for generating guided trajectories. \Cref{fig:scheduler} presents samples from the \acs{ddpm} algorithm \cite{ho2020denoising}, where the adapters were trained on DDIM trajectories \citep{song2020denoising}. \ac{agd} is able to produce high-quality images even when a different scheduler is used during inference. Moreover, \Cref{tab:scheduler} shows that the \ac{fid} scores remain comparable for both schedulers.
\SamplesScheduler
\SDScheduler
\section{Conclusion} \label{sec:conclusion}

This paper introduced \acf{agd}, an efficient approach to achieving the benefits of \acl{cfg} at half the sampling cost. By training lightweight adapters to estimate guided outputs and training on \ac{cfg}-guided trajectories, we address both the computational overhead and the train-inference mismatch of prior guidance distillation methods. Through extensive experiments, we showed that \ac{agd} matches or surpasses \ac{cfg} performance, remains robust to previously unseen guidance scales, and can be trained on a single consumer GPU even for large models such as \ac{sdxl}. Thus, we believe that \ac{agd} offers an efficient and flexible alternative to prior guidance distillation methods while eliminating the sampling overhead of \acl{cfg}. Future research directions could explore integrating \ac{agd} with enhanced guidance algorithms \citep{kynkaanniemi2024applying,karras2024guiding,sadat2024eliminating} and leveraging adapters for other distillation techniques, \eg, adversarial distillation \citep{sauer2024fast,sauer2025adversarial}, to further reduce the sampling time of diffusion models.
{
    \clearpage
    \small
    \bibliographystyle{ieeenat_fullname}
    \bibliography{main}
}

\clearpage
\appendix

\section{Broader impact} \label{app:impact}

Our method accelerates guided sampling in diffusion models, broadening accessibility to large-scale text-to-image or class-conditional generative systems. This can reduce energy consumption and computational barriers to use AI-generated content for various creative applications. However, while advancements in AI-generated content have the potential to improve efficiency and stimulate creativity, it is essential to consider the associated ethical implications. For a more in-depth exploration of ethics and creativity in computer vision, we refer readers to \citet{rostamzadeh2021ethics}

\section{Ablation studies} \label{app:ablations}
This section presents our ablation studies. Unless otherwise specified, all experiments are conducted using the \ac{dit} model for class-conditional generation. We use \ac{fid} as the primary metric to determine the adapter configuration used in the main experiments.

\paragraph{Adapter architecture}
We first examine various design choices for the adapter architecture $\adapter$. Let $\mat{C} = [\vec{c}_1, \ldots, \vec{c}_C]$ represent the matrix containing all conditioning embeddings. The cross-attention and offset adapter architectures are formalized in \Cref{eq:ca,eq:offset} respectively. We further experimented with a gating architecture defined as
\begin{equation}
    \adapter = \lft( \sigma(\tilde{\mat{Z}} \vec{v}) \odot \mathrm{MLP}(\tilde{\mat{Z}}) \rgt) \mat{W},
\end{equation}
where $\tilde{\vec{z}}_j = \lft[ \vec{z}_j, \sum_{i=1}^C \vec{c}_i \rgt]$, $\sigma$ is the sigmoid function, and $\odot: \R^{T} \times \R^{T \times d} \to \R^{T \times d}$ scales each $d$-dimensional vector independently. Lastly, we also considered a positional encoding  adapter architecture given by
\begin{equation}
    \adapter = \mathrm{MLP}(\tilde{\mat{c}}),
\end{equation}
where $\tilde{\vec{c}}_j = \lft[ \vec{e}_j, \sum_{i=1}^C \vec{c}_i \rgt]$ and $\vec{e}_j$ encodes the $j$-th attention time step. Specifically, $\vec{e}_j$ is computed by a Fourier feature encoder \citep{tancik2020fourier}, followed by an \ac{mlp}. The performance of these architectures using the \ac{dit} model are given in \Cref{tab:arch}. Note that for the \ac{dit} model, the offset architecture works the best. However, as shown in \Cref{tab:sd2_arch}, the cross-attention adapter works better for text-to-image models such as Stable Diffusion \citep{rombach2022high}. Hence, we used the offset architecture for class-conditional generation, and the cross-attention adapter for more complex text-to-image models. We also experimented with using dropout in the offset \acp{mlp} for further regularization but found that the model performs best without using any dropout (see \Cref{tab:dropout}). 

\DiTArch
\SDArch
\DiTDropout

\paragraph{Dimensionality of the adapter}
We now examine the impact of adapter dimensionality in \Cref{tab:hidden}. Our results show that increasing the hidden dimension initially improves \ac{fid} but eventually leads to degradation, likely due to overfitting. Therefore, we recommend designing adapters with fewer than 5\% additional parameters w.r.t. the base model.
\DiTHidden

\paragraph{Adapter initialization}
While adapters are typically initialized with zero values such that $\vec{\epsilon}_{[\vec{\theta}, \vec{\psi}]} = \vec{\epsilon}_{\vec{\theta}}$ at initialization \citep{houlsby2019parameter}, 
\Cref{tab:init} shows that Xavier initialization yields better results for guidance distillation. Therefore, we recommend avoiding zero initialization of the adapters for \ac{agd}.
\DiTInit

\paragraph{Training loss functions}
We also explored various loss functions for training \ac{agd}. Specifically, we experimented with $\ell_1(\vec{x}, \vec{y}) = \| \vec{x} - \vec{y} \|_1$ and a weighted $\ell_2(\vec{x}, \vec{y}) = \lambda(t) \| \vec{x} - \vec{y} \|^2_2$, where $\lambda(t)$ is a weighting function depending on the time step. As shown in \Cref{tab:loss}, the simple $\ell_2$ loss with $\lambda(t) = 1$ performs best.
\DiTLoss

\section{Details of the evaluation samples} \label{app:prompts}
\paragraph{\Acs{dit}}
The samples used guidance scale $4$.
\paragraph{\Acs{sd2}}
The samples used guidance scale $10$. From left to right, the prompts used in \Cref{fig:samples:sd2} were:
\begin{enumerate}
    \item ``A cat on the flower.''
    \item ``A close-up of a blooming flower.''
    \item ``A quiet beach at sunset with gentle waves.''
    \item ``A calm lake reflecting the blue sky.''
\end{enumerate}

\paragraph{\Acs{sdxl}}
The samples used guidance scale $12$. Further, the prompts used in \Cref{fig:samples:sdxl} were:
\begin{enumerate}
    \item ``A modern reinterpretation of a classical Renaissance painting, where futuristic elements and digital motifs merge with traditional portraiture.''
    \item ``A fantastical scene of a celestial garden floating in space, featuring luminous, otherworldly flora against a backdrop of swirling galaxies.''
    \item ``A hyper-realistic digital painting of a futuristic metropolis at sunset, with neon lights reflecting off rain-soaked streets and towering holograms.''
    \item ``A cozy winter scene of a remote mountain village, with softly glowing windows, snow-covered rooftops, and a star-filled night sky.''
\end{enumerate}

\section{Additional visual samples} \label{app:uncurated}
\Cref{fig:uncurated:dit:207,fig:uncurated:dit:812,fig:uncurated:dit:88,fig:uncurated:dit:279,fig:uncurated:dit:537,fig:uncurated:dit:387,fig:uncurated:sd2:flowers,fig:uncurated:sd2:teddy,fig:uncurated:sd2:kites,fig:uncurated:sdxl:stuffed,fig:uncurated:sdxl:capybara,fig:uncurated:sdxl:dragon} provides more uncurated samples of \ac{agd} and \ac{cfg} on various models used in the paper. The samples are best viewed when zoomed in.

\section{Additional implementation details}
\label{app:details}

\Cref{sec:adapters} provides the main implementation details. The \ac{dit} model was trained with a batch size of 64 for 5000 gradient steps, the \ac{sd2} model with a batch size of 8 for 5000 gradient steps, and the \ac{sdxl} model with a batch size of 1 for 20000 gradient steps. These settings were selected based on the maximum batch size that fit within 24,GB of VRAM. For all quantitative experiments, we set the guidance scale to the value that achieved the best \ac{fid} for each method. The AGD implementation will be publicly released to support further research on guidance distillation.

The \ac{fid} scores for class-conditional models were computed using 10k generated samples and the entire ImageNet validation set. For text-to-image models, we used the full COCO-2017 validation set as the real data. All metrics were computed using the ADM evaluation code base \citep{dhariwal2021diffusion} to ensure fairness across experiments.

\NewDocumentCommand{\UncuratedSamples}{mmm}{
\begin{figure*}[ht]
    \centering
    \begin{tikzpicture}[every node/.style={inner sep=0pt, outer sep=0pt}]
        \node (img1) {\includegraphics[width=0.48\linewidth]{figures/samples/uncurated/#1/#2/cfg}};
        \node [right=0.4 of img1] (img2) {\includegraphics[width=0.48\linewidth]{figures/samples/uncurated/#1/#2/agd}};
        \node [above=0.2 of img1] {CFG};
        \node [above=0.2 of img2] {AGD (Ours)};
    \end{tikzpicture}
    \caption{#3}
    \label{fig:uncurated:#1:#2}
\end{figure*}
}

\UncuratedSamples{dit}{812}{Uncurated samples using \ac{dit}. Class label: ``Space shuttle'' (812), guidance scale: $2$.}

\UncuratedSamples{dit}{207}{Uncurated samples using \ac{dit}. Class label: ``Golden retriever'' (207),  guidance scale: $3$.}

\UncuratedSamples{dit}{88}{Uncurated samples using \ac{dit}. Class lable: ``Macaw'' (88), guidance scale: $3$.}

\UncuratedSamples{dit}{279}{Uncurated samples using \ac{dit}. Class label: ``Arctic fox'' (279), guidance scale: $4$.}

\UncuratedSamples{dit}{387}{Uncurated samples using \ac{dit}. Class lable: ``Red panda'' (387), guidance scale: $5$.}

\UncuratedSamples{dit}{537}{Uncurated samples using \ac{dit}. Class label: ``Dog sled'' (537), guidance scale: $6$.}

\UncuratedSamples{sd2}{flowers}{Uncurated samples using \ac{sd2}. Prompt: ``A close up of a clear vase with flowers.'', guidance scale: $10$.}

\UncuratedSamples{sd2}{teddy}{Uncurated samples using \ac{sd2}. Prompt: ``A set of plush toy teddy bears sitting in a sled.'', guidance scale: $10$.}

\UncuratedSamples{sd2}{kites}{Uncurated samples using \ac{sd2}. Prompt: ``People flying kites in a park on a windy day.'', guidance scale: $10$.}

\UncuratedSamples{sdxl}{stuffed}{Uncurated samples using \ac{sdxl}. Prompt: ``Two stuffed animals posed together in black and white.'' guidance scale: $12$.}

\UncuratedSamples{sdxl}{capybara}{Uncurated samples using \ac{sdxl} Prompt: ``A capybara made of lego sitting in a realistic, natural field.'', guidance scale: $12$.}

\UncuratedSamples{sdxl}{dragon}{Uncurated samples using \ac{sdxl} Prompt: ``A close-up of a fire spitting dragon, cinematic shot.'', guidance scale: $12$.}


\end{document}